\documentclass[12pt]{article} 
\usepackage[sectionbib]{natbib}
\usepackage{array,epsfig,fancyheadings,rotating}
\usepackage[]{hyperref}  
\usepackage{sectsty, secdot}
\sectionfont{\fontsize{12}{14pt plus.8pt minus .6pt}\selectfont}
\renewcommand{\theequation}{\thesection\arabic{equation}}
\subsectionfont{\fontsize{12}{14pt plus.8pt minus .6pt}\selectfont}

\usepackage{amsmath}
\usepackage{amssymb}
\usepackage{amsfonts}
\usepackage{multirow}
\usepackage{amsthm}
\usepackage{bbm}
\usepackage{enumerate}
\usepackage[ruled,vlined]{algorithm2e}
\usepackage{setspace}

\usepackage{gensymb}
\graphicspath{{Figures/}}
\usepackage{hyperref}
\usepackage{epsfig, epsf, color}

\theoremstyle{definition}

\begin{document}


\renewcommand{\baselinestretch}{2}
\setlength\abovedisplayskip{2pt}
\setlength\belowdisplayskip{2pt}

\markright{ \hbox{\footnotesize\rm Statistica Sinica, to appear
}\hfill\\[-13pt]
\hbox{\footnotesize\rm
}\hfill }

\markboth{\hfill{\footnotesize\rm Wanfang Chen, Yuxiao Li, Brian J Reich and Ying Sun} \hfill}
{\hfill {\footnotesize\rm DeepKriging} \hfill}

\renewcommand{\thefootnote}{}
$\ $\par


\fontsize{12}{14pt plus.8pt minus .6pt}\selectfont \vspace{0.8pc}
\centerline{\large\bf DeepKriging: Spatially Dependent Deep Neural Networks}
\vspace{2pt} 
\centerline{\large\bf for Spatial Prediction}
\vspace{.4cm} 
\centerline{Wanfang Chen$^{1}$, Yuxiao Li$^{2}$, Brian J Reich$^{3}$ and Ying Sun$^{2}$} 
\begin{center}
\it $^{1}$ East China Normal University\\ 
$^{2}$ King Abdullah University of Science and Technology\\ 
$^{3}$ North Carolina State University
\end{center}
 \fontsize{9}{11.5pt plus.8pt minus.6pt}\selectfont


\begin{quotation}
\noindent {\it Abstract:}
In spatial statistics, a common objective is to predict values of a spatial process at unobserved locations by exploiting spatial dependence. Kriging provides the best linear unbiased predictor using covariance functions and is often associated with Gaussian processes. However, when considering non-linear prediction for non-Gaussian and categorical data, the Kriging prediction is no longer optimal, and the associated variance is often overly optimistic. Although deep neural networks (DNNs) are widely used for general classification and prediction, they have not been studied thoroughly for data with spatial dependence.  In this work, we propose a novel DNN structure for spatial prediction, where the spatial dependence is captured by adding an embedding layer of spatial coordinates with basis functions. We show in theory and simulation studies that the proposed DeepKriging method has a direct link to Kriging in the Gaussian case, and it has multiple advantages over Kriging for non-Gaussian and non-stationary data, i.e., it provides non-linear predictions and thus has smaller approximation errors, it does not require operations on covariance matrices and thus is scalable for large datasets, and with sufficiently many hidden neurons, it provides the optimal prediction in terms of model capacity. We further explore the possibility of quantifying prediction uncertainties based on density prediction without assuming any data distribution. Finally, we apply the method to predicting PM$_{2.5}$ concentrations across the continental United States.

\vspace{9pt}
\par
\end{quotation}\par

\def\thefigure{\arabic{figure}}
\def\thetable{\arabic{table}}

\renewcommand{\theequation}{\thesection.\arabic{equation}}

\fontsize{12}{14pt plus.8pt minus .6pt}\selectfont

\section{Introduction}
\label{s:intro}

Spatial prediction is at the heart of spatial and spatio-temporal statistics. It is aimed at predicting values of a spatial process at unobserved locations by accounting for the spatial dependence in the region of interest.  Traditional applications of spatial prediction are in the fields of geological and environmental science \citep{cressie2015statistics}, and they have been extended to other fields, such as biological sciences, computer vision, economics and public health \citep{anselin2001spatial, austin2002spatial,  waller2004applied, franchi2018supervised}. 

The primary collection of spatial prediction methods are based on the best linear unbiased prediction (BLUP), also referred to as Kriging \citep{matheron1963principles}. Kriging prediction is a weighted average of observed data, where the weights are determined by the spatial covariance function or variogram of the random process. Under the Gaussian assumption, Kriging also provides the full predictive distribution. { Applying Kriging requires estimating the spatial covariance function, which is commonly assumed to be stationary.} However, physical processes tend to be non-Gaussian and non-stationary. For instance, the data on wind speed and fine particles (PM$_{2.5}$) exposures are positive, right-skewed, and sometimes heavy-tailed \citep{hennessey1977some, adgate2002spatial}, and the spatial covariance typically varies across space, e.g., in urban versus rural areas \citep{sampson2013regionalized}.  It is possible to derive the best linear prediction for certain parametric non-Gaussian processes \citep{xu2017tukey, rimstad2014skew} and certain non-stationary covariance structures \citep{fuentes2002spectral, paciorek2004nonstationary, Yuxiao2018}, but Kriging for more general spatial processes remains an open problem. Another drawback of Kriging is that it is computationally prohibitive for large spatial datasets, since it involves computing the inversion of an $N\times N$ covariance matrix, where $N$ is the number of observed locations \citep{heaton2019case}, and the computation requires $O(N^3)$ time and $O(N^2)$ memory complexity based on the typical Cholesky decomposition approach. 

Recently, deep learning or deep neural networks (DNNs) have become the most powerful prediction tools for a wide range of applications, especially in computer vision and natural language processing \citep{lecun2015deep}. DNNs are effective for predictions with complex features such as non-linearity and non-stationarity, and they are computationally efficient in analyzing massive datasets using GPUs \citep{najafabadi2015deep}. Therefore, it would be promising to apply DNNs to spatial predictions. However, classical DNNs cannot incorporate the spatial dependence appropriately. Applications in spatial prediction with neural networks usually simply include spatial coordinates as features \citep{cracknell2014geological}, which may not be sufficient. Recently, convolutional neural networks (CNNs, \citealt{krizhevsky2012imagenet}) have been claimed to successfully capture the spatial and temporal dependencies in image processing through the relevant filters. However, the framework is designed for applications with a large feature space, and often requires large training labels as the ground truth, which does not fit for many spatial prediction problems, where only in-situ and sparse observations are available. 

To tackle the above-mentioned problems, we develop an effective deep neural network for spatial prediction that\\
 1) builds a direct link between DNNs and Kriging in spatial prediction;\\
 2) models the spatial dependence through a set of basis functions;\\
 3) does not require matrix operations and is scalable for large datasets;\\
 4) provides a non-linear predictor in covariates or generally in observations;\\
 5) has a Gaussian process representation, and brings more flexible spatial covariance structures than simply using the coordinates as features; \\
 6) suits for different data types, e.g., non-Gaussian or non-stationary data; \\
 7) potentially measures the uncertainty through predictive density functions without assuming any data distribution.

We call our method ``DeepKriging'' with the aim of achieving the optimal spatial prediction, similar to the original use of Kriging \citep{cressie1990origins}, but by using deep neural networks. 
We also conduct simulation studies and apply our approach to the PM$_{2.5}$ concentration data across the continental United States to show the performance of DeepKriging compared to Kriging and other naive DNN methods. The rest of our paper is organized as follows.  Section~2 introduces the construction of our DeepKriging method. Section~3 provides its theoretical properties. Section~4 presents some simulation studies to show the performance of DeepKriging.  
Section~5 applies DeepKriging to predict PM$_{2.5}$ concentration in the U.S. Section 6 summarizes the main results and suggests directions for future work. 

\section{Methodology}\label{sec:deepkrigng}
\subsection{Deep learning in spatial prediction}\label{sec:DK}
Suppose $\mathbf{z} = \{z(\mathbf{s}_1),\dots, z(\mathbf{s}_N)\}^T$ are measurements observed at $N$ spatial locations from a real-valued spatial process $\{Y(\mathbf{s}): \mathbf{s} \in D\}, \;D \subseteq \mathbb{R}^d$. The goal of spatial prediction is to find the optimal predictor $\widehat{Y}^{\mathrm{opt}}(\mathbf{s}_0)$ of the true process at an unobserved location $\mathbf{s}_0$, as a function of $\mathbf{z}$. In decision theory, $\widehat{Y}^{\mathrm{opt}}(\mathbf{s}_0)$ is the minimizer of an expected loss function or risk function \citep{degroot2005optimal}. That is,
\begin{equation}
 \widehat{Y}^{\mathrm{opt}}(\mathbf{s}_0) = \underset{\hat{Y}}{\mathrm{argmin}}
 \;\mathbb{E}\{L(\widehat{Y}(\mathbf{s}_0),Y(\mathbf{s}_0))\} =  \underset{\hat{Y}}{\mathrm{argmin}}
 \; R(\widehat{Y}(\mathbf{s}_0),Y(\mathbf{s}_0)),
 \label{eq:loss}
\end{equation}
where $L(\cdot,\cdot)$ is a loss function and $R(\cdot,\cdot)$ is a risk function. Under the mean squared error (MSE) loss, the optimal predictor is $\widehat{Y}^{\mathrm{opt}}(\mathbf{s}_0)= \mathbb{E}\{Y(\mathbf{s}_0)|\mathbf{z}\}$ if it is finite. This predictor has multiple good properties such as unbiasedness and asymptotic normality under regularity assumptions \citep{lehmann2006theory}. In particular, if $Y(\mathbf{s}_0)$ and $\mathbf{z}$ are jointly Gaussian, the conditional mean is a linear combination of $\mathbf{z}$; if $Y(\mathbf{s}_0)$ and $\mathbf{z}$ are not jointly Gaussian, the conditional mean obtained with Gaussian assumption remains the best linear unbiased prediction (BLUP), which is called Kriging. However, as mentioned before, the Kriging predictor is sub-optimal for non-Gaussian data, and it is not scalable for large data size.

In this work, we use deep learning to approximate the optimal predictor $\widehat{Y}^{\mathrm{opt}}(\mathbf{s}_0)$ in \eqref{eq:loss} by the output of the neural network. The optimal neural network predictor is given by $f_{\mathrm{NN}}^{\mathrm{opt}}(\mathbf{s}_0)=\mathrm{argmin}_{f_{\mathrm{NN}}} R\{f_{\mathrm{NN}}(\mathbf{s}_0),Y(\mathbf{s}_0)\}$,
where $f_{\mathrm{NN}}(\cdot)\in \mathcal F$ can be any function in the function space $\mathcal F$ expressible by a family of neural networks, and $f_{\mathrm{NN}}^{\mathrm{opt}}(\cdot)$ is the best function in $\mathcal F$ in terms of minimizing a certain risk $R(\cdot,\cdot)$. The inputs of the neural network can be relevant covariates $\mathbf{x}(\mathbf{s}_0)$ and other features at $\mathbf{s}_0$. Typically, we write $f_{\mathrm{NN}}(\mathbf{s};\boldsymbol{\theta})$ as a parametric model with unknown parameters $\boldsymbol{\theta}$, which include the weights and biases in the neural network. Note that the optimal neural network predictor $f_{\mathrm{NN}}^{\mathrm{opt}}(\mathbf{s}_0)$ is practically unreachable since $Y(\mathbf{s}_0)$ is unknown. In practice, we approximate the predictor by minimizing the empirical loss function over the training set $\mathbf{z}$ \citep{goodfellow2016deep}; i.e., the final predictor is $\widehat{Y}_{\mathrm{NN}}(\mathbf{s}_0)= f_{\mathrm{NN}}(\mathbf{s}_0;\widehat{\boldsymbol{\theta}})$, with 
\begin{equation}
\widehat{\boldsymbol{\theta}} = \underset{\boldsymbol{\theta}}{\mathrm{argmin}}
\;\frac{1}{N}\sum_{n=1}^NL\{f_{\mathrm{NN}}(\mathbf{s}_n;\boldsymbol{\theta}),z(\mathbf{s}_n)\}.
\label{eq:theta}
\end{equation}

Applying this framework of classical neural network directly to spatial prediction is problematic in at least two aspects:  classical DNNs does not account for the spatial dependence, and spatial prediction typically has limited observed features rather than excessive features in common applications of neural networks. In particular, assume that the spatial process $Y(\mathbf{s})$ is modeled by $Y(\mathbf{s}) = \mathbf{x}(\mathbf{s})^T\boldsymbol{\beta} + \nu(\mathbf{s})$, where $\mathbf{x}(\mathbf{s})\in \mathbb{R}^P$ is a vector process of $P$ known covariates, $\boldsymbol{\beta}$ is a vector of coefficients, and $\nu(\mathbf{s})$ is a spatially dependent and zero-mean random process with a generally non-stationary covariance function: $\mbox{Cov}(\nu(\mathbf{s}),\nu(\mathbf{s'}))=\rm{C}(\mathbf{s},\mathbf{s'})$. In neural networks, we usually assume that $Y(\mathbf{s})$ are mutually independent conditional on the features $\mathbf{x}(\mathbf{s})$. However, this assumption is not reasonable in spatial prediction because the covariates $\mathbf{x}(\mathbf{s})$ only contribute to the mean structure of $Y(\mathbf{s})$ and $\nu(\mathbf{s})$ remains a spatially correlated process. Hence, more features apart from $\mathbf{x}(\mathbf{s})$ are needed to model the spatial dependence in applying the neural networks.

To account for the spatial information, the most natural way is to add $d$ coordinates (e.g., longitude and latitude) to the features, in the hope that the neural networks can learn the dependent term $\nu(\mathbf{s})$ as a function of $\mathbf{s}$ \citep{cracknell2014geological}. By doing that, the adjusted features become $\mathbf{x}^{adj}(\mathbf{s})=(\mathbf{x}(\mathbf{s})^T,\mathbf{s})^T$. However, this does not help much in enlarging the feature space since usually the dimension of coordinates has $d\leq 3$. Moreover, the associated neural network may not be efficient, since if the true function is far from linear, it may take huge effort for the neural network to achieve a good approximation. For instance, the optimal predictor under the Gaussian assumption and MSE loss is the Kriging predictor, which is linear in $\mathbf{x}(\mathbf{s})$ but obviously non-linear in coordinates $\mathbf{s}$; this is a special case where the natural structure of neural networks may not work. 

Going deeper into the form of Kriging prediction may give us a hint about the appropriate way to incorporate the spatial dependence in the DNN. Suppose $\mathbf{z}$ is observed from a generalized additive model: $Z(\mathbf{s})= Y(\mathbf{s})+\varepsilon(\mathbf{s})$, where $Y(\mathbf{s}) = \mathbf{x}(\mathbf{s})^T\boldsymbol{\beta} + \nu(\mathbf{s})$ as defined above, and $\varepsilon(\mathbf{s})$ is a white noise process, called the nugget effect, with zero mean and variance $\sigma^2(\mathbf{s})$, caused by measurement inaccuracy and fine-scale variability.  The (universal) Kriging prediction is
\begin{equation}
\widehat{Y}_{\mathrm{UK}}(\mathbf{s}_0) = \mathbf{x}(\mathbf{s}_0)^T\widehat{\boldsymbol{\beta}} + \mathbf{c}(\mathbf{s}_0)^T\boldsymbol{\Sigma}^{-1}(\mathbf{z}-\mathbf{X}\widehat{\boldsymbol{\beta}}),
\label{eq:Kriging}
\end{equation}
where $\mathbf{X}=(\mathbf{x}(\mathbf{s}_1),\dots,\mathbf{x}(\mathbf{s}_N))^T$ is an $N\times P$ matrix, $\mathbf{c}(\mathbf{s}_0)=\mbox{Cov}(\mathbf{Z},Z(\mathbf{s}_0))$, $\boldsymbol{\Sigma}=\mbox{Cov}(\mathbf{Z},\mathbf{Z}^T)$, and $\widehat{\boldsymbol{\beta}}=(\mathbf{X}^T\boldsymbol{\Sigma}^{-1} \mathbf{X})^{-1}\mathbf{X}^T\boldsymbol{\Sigma}^{-1}\mathbf{z}$. The spatial dependence is incorporated in $\widehat{Y}_{\mathrm{UK}}(\mathbf{s}_0)$ via a linear function of the covariance vector $\mathbf{c}(\mathbf{s}_0)$, but its coefficient $\boldsymbol{\Sigma}^{-1}(\mathbf{z}-\mathbf{X}\widehat{\boldsymbol{\beta}})$ is unknown. This motivates us to use a set of known nonlinear functions as the embedding of $\mathbf{s}$ in the features to characterize the spatial process $\nu(\mathbf{s})$ in the neural network. This can be done by the lights of the Karhunen--Lo{\`e}ve (KL) theorem \citep{adler2010geometry}, which establishes that $\nu(\mathbf{s})$ admits a decomposition $\nu(\mathbf{s})=\sum_{k=1}^{\infty}w_k \phi_k(\mathbf{s})$, where $w_k$'s are pairwise uncorrelated random variables and $\phi_k(\mathbf{s})$'s are pairwise orthogonal basis functions in the domain of $\nu(\mathbf{s})$. Hence, $\nu(\mathbf{s})$ can be linearly quantified by nonlinear basis functions of $\mathbf{s}$. 

In practice, the prediction of $\nu(\mathbf{s})$ is typically the truncated KL expansion based on the property that given any orthonormal basis functions $\phi_k(\mathbf{s})$, we can find some large integer $K$, so that $\nu(\mathbf{s})$ can be approximated by the finite weighted sum of basis functions, i.e.,  $\widehat{\nu}(\mathbf{s})=\sum_{k=1}^{K}w_k \phi_k(\mathbf{s})$. Based on the KL theorem, the form of basis functions is not as important as the number of basis functions to approximate the spatial random effect $\nu(\mathbf{s})$. This can also be supported by the additional simulations we conduct in Section~S4.1 of the Supplementary Material. Multiple types of basis functions can be used, such as the smoothing spline basis functions \citep{wahba1990spline}, the wavelet basis functions \citep{vidakovic2009statistical}, and the radial basis functions \citep{friedman2001elements}. By adding an embedding layer with sufficiently large $K$, the width of the neural network is greatly increased so that the network incorporates more spatial information than using the coordinates alone. A similar idea has been used in the recommendation systems by \citet{cheng2016wide}.

\subsection{DeepKriging: a spatially dependent neural network}\label{sec:DNN}
In this section, we use a simple DNN to illustrate our DeepKriging framework. Our model can be potentially used in other deep learning frameworks such as convolutional neural networks (CNNs) and recurrent neural networks (RNNs). 

First, we need to choose the value for $K$ and basis functions to approximate the spatial process $\nu(\mathbf{s})$. We adopt the idea in \cite{nychka2015multiresolution}, who developed a multi-resolution model for spatial prediction for large datasets. The radial basis functions at each level of resolution are constructed using a Wendland compactly supported correlation function with the nodes arranged on a rectangular grid. In particular, at a certain level of resolution, let $\{\mathbf{u}_j\}$, $j=1,\ldots,m$, be a rectangular grid of points (or node points in the radial basis function terminology) and let $\theta$ be a scale parameter. The basis functions are given by $\phi^{*}_j(\mathbf{s})=\phi(\|\mathbf{s}-\mathbf{u}_j\|/\theta)$, where 
	\begin{equation*}
		\phi(d)=\begin{cases}
			(1-d)^6(35d^2+18d+3)/3, & d\in[0,1] \\
			0,& \mbox{otherwise}.
		\end{cases}
	\end{equation*}
Hence, the embedding layer uses mutual distance locally to each knot location, implying that the spatial patterns are location invariant locally. As a result, the proposed DeepKriging is able to model the spatial non-stationarity; as we will show in Section~3.3, the induced covariance functions of an infinitely wide DeepKriging network are in general non-stationary. The scale parameter $\theta$ is set to be $2.5$ times the associated knots spacing according to \cite{nychka2015multiresolution}. The grid at each finer level increases by a factor of two and the basis functions are scaled to have a constant overlap. In particular, in the $h$-th level, the number of knots is chosen to be $K_h = (9\times 2^{h-1} +1)^d$, where $d$ is the spatial dimension. For a massive dataset and to obtain $K\geq N$, we need $H=1+ \lceil\log_2(\sqrt[d]N/10)\rceil$ levels. Therefore, for a four-level model for instance, we need $K=10+19+37+73=139$ basis functions in one dimensional space and $K=10^2+19^2+37^2+73^2=7159$ basis functions in two dimensional space. This scheme gives a good approximation to standard covariance functions and also has the flexibility to fit more complicated shapes. The approach of multi-resolution approximation for massive spatial datasets has also been adopted in other research works; see \cite{katzfuss2017multi} and the references therein.

Then, for any coordinate $\mathbf{s}$, we compute the $K$ basis functions to get the embedded vectors $\boldsymbol{\phi}(\mathbf{s})=(\phi_1(\mathbf{s}),\dots,\phi_K(\mathbf{s}))^T$. The basis functions are recommended to be orthogonal based on the KL expansion. Then, let $\mathbf{x}_{\phi}(\mathbf{s})=(\mathbf{x}(\mathbf{s})^T, \boldsymbol{\phi}(\mathbf{s})^T)^T$ be the embedded input of length $P+K$, and specify~an $L$-layer DNN as
\begin{equation}
{
\begin{array}{r@{}l}
&\mathbf{u}_1(\mathbf{s})=\mathbf{W}_1\mathbf{x}_{\phi}(\mathbf{s}) + \mathbf{b}_1, \hspace{2mm} \mathbf{a}_1(\mathbf{s})=\psi_1(\mathbf{u}_1(\mathbf{s}));\\
&\mathbf{u}_2(\mathbf{s})=\mathbf{W}_2\mathbf{a}_1(\mathbf{s})+ \mathbf{b}_2, \hspace{2mm} \mathbf{a}_2(\mathbf{s})=\psi_2(\mathbf{u}_2(\mathbf{s}));\\
&\dots\\
&\mathbf{u}_L(\mathbf{s})=\mathbf{W}_{L}\mathbf{a}_{L-1}(\mathbf{s})+ \mathbf{b}_L, \hspace{2mm} f_{\mathrm{DK}}(\mathbf{s})=\psi_L(\mathbf{u}_L(\mathbf{s})).
\end{array}
}
\label{eq:DK}
\end{equation}
For the $l$-th layer with $N_l$ neurons, $\mathbf{W}_l$ is the $N_{l}\times N_{l-1}$ weight matrix, $\mathbf{b}_l$ is the bias vector of length $N_l$, $\mathbf{a}_l$ is the neuron vector of length $N_l$, and $\psi_l(\cdot)$ is the activation function. The output of this neural network is $f_{\mathrm{DK}}(\mathbf{s})$, which is a function of the weights and biases. Let $\boldsymbol{\theta}$ be the vector of unknown weights and biases, and $\widehat{\boldsymbol{\theta}}$ be the estimate via Equation \eqref{eq:theta} based on the training sample. The final DeepKriging prediction at an unobserved location $\mathbf{s}_0$ is defined as $\widehat{Y}_{\rm{DK}}(\mathbf{s}_0)= f_{\rm{DK}}(\mathbf{s}_0;\widehat{\boldsymbol{\theta}})$.

\begin{figure}[!h] 
 \centering
\includegraphics[width=0.85\textwidth]{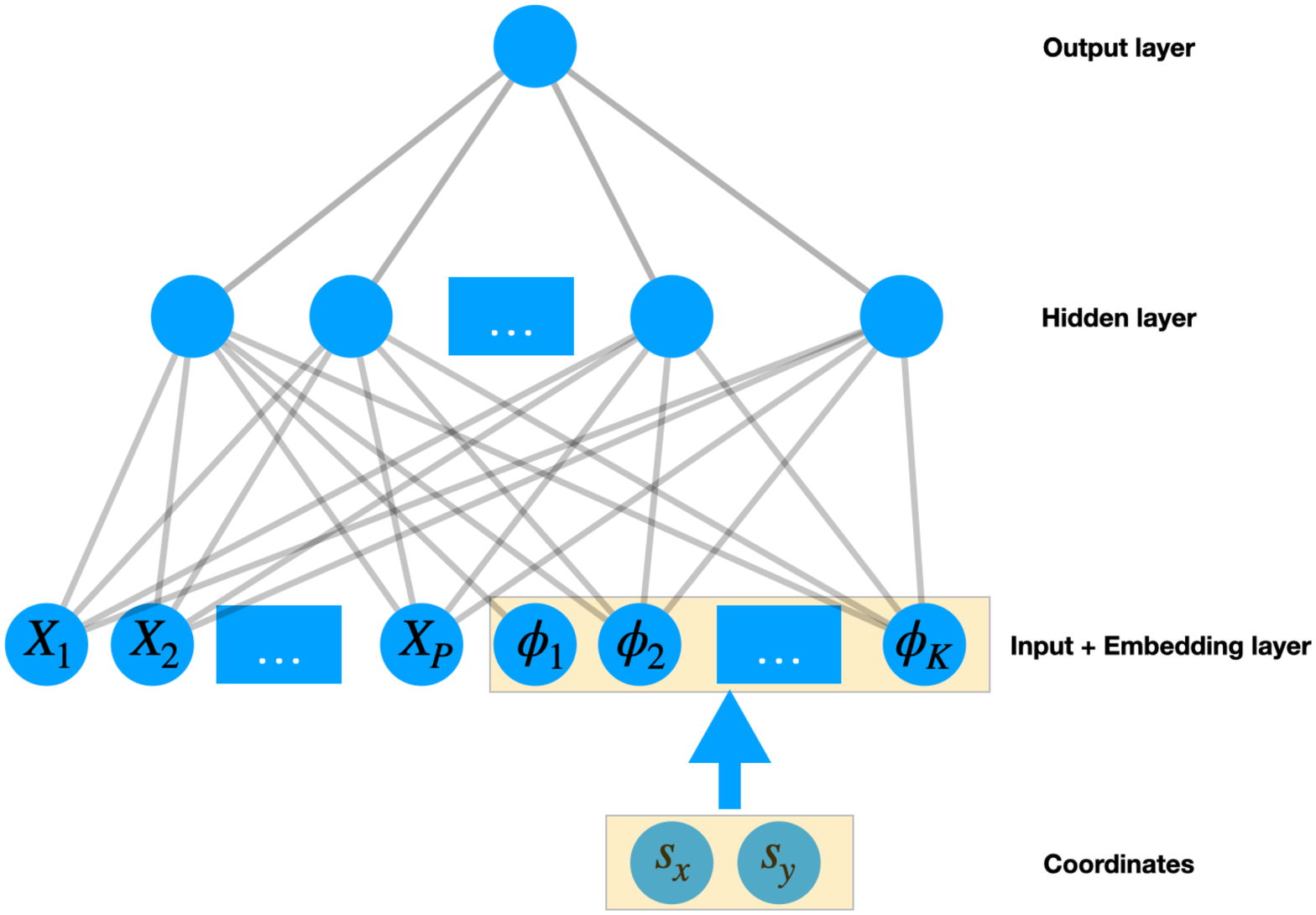}
\vspace*{-.5cm}
\caption{\baselineskip=20pt Visualization of the DeepKriging structure in 2D spatial prediction based on a three-layer DNN}
\label{fig:ANN}
\end{figure}

One major advantage of our DeepKriging method is that we can adjust the number of neurons, activation functions and loss functions to fit for different data types and model interpretations. For example, for predicting continuous variables as in a regression problem, we choose $N_L=1$, $\psi_L(\cdot)$ to be an identity function, and the loss function to be the MSE. Figure~\ref{fig:ANN} provides a visualization of a DeepKriging structure in two dimensional prediction for continuous data. For predicting categorical variables as in a classification problem, we choose $N_L$ to be the number of categories, $\psi_L(\cdot)$ to be a softmax function, and the loss function to be the cross entropy loss. For the activation functions in the hidden layers, we choose the rectified linear unit (ReLU) in default, which allows us to keep the linear relationship in the KL expansion but add some deactivated neurons to select the best number of basis functions. 
The DeepKriging structure also allows the covariate effects to be spatially varying.
 
Regularization of the DeepKriging network structure includes adding dropout layers to mitigate overfitting, adding batch-normalization layers to regularize the covariates and basis functions to the same scale, and removing all-zero columns in the basis matrix whenever they are present due to the compactly supported structure of the basis function. Details of the default setting of our DeepKriging network structure are included in Section~S2 of the Supplementary Materials. The time complexity of our DeepKriging method is about $O(N_{\rm{neuron}})$, where $N_{\rm{neuron}}$ is the number of neurons in the network. The computation cost depends on the width and depth of the network, and the computation is highly parallelizable and can be largely accelerated by CPUs and GPUs. 

\section{Theoretical Properties of DeepKriging}\label{sec:theory}
DeepKriging provides a novel spatial prediction framework using deep learning.  It differs from classical Kriging methods in several aspects. First, Kriging prediction is a linear combination of observations; in contrast, DeepKriging prediction is linked to the observations via the weights and biases through model training and is typically nonlinear in observations (see Section~S3.1). Second, DeepKriging does not assume a Gaussian process with a certain covariance function but models spatial dependence by basis functions. Last, unlike Kriging which predicts the random process $Y(\mathbf{s})$ at an unobserved location, DeepKriging approximates the process using a deterministic continuous function.

In this section, we provide important theoretical properties of DeepKriging including 1) the underlying relationship between DeepKriging and Kriging; 2) how accurate Deepkriging can be in terms of the prediction error compared to Kriging; and 3) how the spatial dependence is measured in the DeepKriging framework. These three aspects are critical for understanding our DeepKriging method, and will be illustrated in the following subsections, respectively.
 
\subsection{The link between DeepKriging and Kriging-based methods}
DeepKriging is closely related to Kriging and its associated variants, which can be classified as multi-resolution processes \citep{nychka2015multiresolution, kleiber2015equivalent, katzfuss2017multi} and Gaussian predictive processes \citep{banerjee2008gaussian, banerjee2010hierarchical}, all leading to spatial predictions that can be treated as linear functions of embedded features $\mathbf{x}_{\phi}(\mathbf{s}_0)$, and thus can be potentially approximated by DeepKriging. 

One example is the fixed rank Kriging (FRK) proposed by \citet{cressie2008fixed}, who used one of the low-rank approximations of the covariance matrix in order to speed up the computation of universal Kriging. Similar to DeepKriging, they represent the spatial random effects $\nu(\mathbf{s})$ by $K$ basis functions, i.e., $\nu(\mathbf{s})=\boldsymbol{\phi}(\mathbf{s})^T\boldsymbol{\eta}$, where $\boldsymbol{\eta}$ is a $K$ dimensional Gaussian random vector with $\mathbb{C}\mathrm{ov}(\boldsymbol{\eta})=\boldsymbol{\Sigma}_K$. They also assume that the model for $Y(\mathbf{s})$ is $Y(\mathbf{s}) = \mathbf{x}(\mathbf{s})^T\boldsymbol{\beta} + \nu(\mathbf{s})= \mathbf{x}(\mathbf{s})^T\boldsymbol{\beta}+\boldsymbol{\phi}(\mathbf{s})^T\boldsymbol{\eta}$. The covariance matrix of $Z(\mathbf{s})=Y(\mathbf{s})+\varepsilon(\mathbf{s})$, where $\varepsilon(\mathbf{s})$ is a white noise with variance $\sigma^{2}(\mathbf{s})$, is given by $\boldsymbol{\Sigma}=\boldsymbol{\Phi}\boldsymbol{\Sigma}_K\boldsymbol{\Phi}^T+\boldsymbol{V}$, where $\boldsymbol{\Phi}=\{\boldsymbol{\phi}(\mathbf{s}_1),\dots,\boldsymbol{\phi}(\mathbf{s}_N)\}^T$ is an $N\times K$ basis matrix and $\boldsymbol{V}=\mathrm{diag}\{\sigma^2(\mathbf{s}_1), \dots,\sigma^2(\mathbf{s}_N)\}$ is an $N\times N$ diagonal matrix. The FRK prediction as a linear function of $\mathbf{z}$ is given by
\begin{equation}
\widehat{Y}_{\mathrm{FRK}}(\mathbf{s}_0) = \mathbf{x}(\mathbf{s}_0)^T\widehat{\boldsymbol{\beta}} + \boldsymbol{\phi}(\mathbf{s}_0)^T\boldsymbol{\Sigma}_K\boldsymbol{\Phi}^T\boldsymbol{\Sigma}^{-1}(\mathbf{z}-\mathbf{X}\widehat{\boldsymbol{\beta}}),
\label{eq:FRK}
\end{equation}
where $\mathbf{X}=(\mathbf{x}(\mathbf{s}_1),\cdots,\mathbf{x}(\mathbf{s}_N))^T$ is an $N\times P$ matrix, $\widehat{\boldsymbol{\beta}}=(\mathbf{X}^T\boldsymbol{\Sigma}^{-1} \mathbf{X})^{-1}\mathbf{X}^T\boldsymbol{\Sigma}^{-1}\mathbf{z}$, and $\mathbf{\Sigma}^{-1}$ has a computationally simple form which involves inverting the fixed rank $K\times K$ positive definite matrix $\mathbf{\Sigma}_{K}$ and the $N\times N$ diagonal matrix $\boldsymbol{V}$.
Writing Equation \eqref{eq:FRK} as $\widehat{Y}_{\mathrm{FRK}}(\mathbf{s}_0) =\mathbf{x}(\mathbf{s}_0)^T\widehat{\boldsymbol{\beta}} + \boldsymbol{\phi}(\mathbf{s}_0)^T\widehat{\boldsymbol{\alpha}},\mbox{ where } \widehat{\boldsymbol{\alpha}}=\boldsymbol{\Sigma}_K\boldsymbol{\Phi}^T\boldsymbol{\Sigma}^{-1}(\mathbf{z}-\mathbf{X}\widehat{\boldsymbol{\beta}})$,
implies that the FRK prediction $\widehat{Y}_{\mathrm{FRK}}(\mathbf{s}_0)$ is linear in $P$ covariates $\mathbf{x}(\mathbf{s}_0)$ and $K$ basis functions $\boldsymbol{\phi}(\mathbf{s}_0)$. This is a special case of DeepKriging when we set all of the activation functions to be linear.

FRK usually chooses $K$ to be much smaller than $N$ in order to speed up the computation for large datasets. Since the covariance $\boldsymbol{\Phi}\boldsymbol{\Sigma}_K\boldsymbol{\Phi}^T$ has at most rank $K$, such a low-rank approximation of the covariance matrix may fail to capture the high-frequency variation or small-scale spatial dependence in the spatial process \citep{stein2014limitations}. In contrast, for DeepKriging, $K$ needs to be sufficiently large ($K>N$) in order to have a good approximation of the spatial random effect $\nu(\mathbf{s})$, so that our method captures more spatial information in the prediction.

By setting $K=N$ in the FRK, we can see that the (universal) Kriging prediction in Equation \eqref{eq:Kriging} is also a linear function of $\mathbf{x}_{\phi}(\mathbf{s}_0)=(\mathbf{x}(\mathbf{s}_0)^T, \boldsymbol{\phi}(\mathbf{s}_0)^T)^T$. A detailed proof is provided in Section~S1.1 in the Supplementary Materials. This result implies that the Kriging prediction with any covariance function can be linearly expressed by the embedding features $\mathbf{x}_{\phi}(\mathbf{s}_0)$. In this sense, DeepKriging generalizes Kriging by allowing for nonlinear functions of $\mathbf{x}_{\phi}(\mathbf{s}_0)$ in the prediction.

\subsection{DeepKriging in decision theory}
Our DeepKriging prediction procedure conventionally follows an approximation-estimation decomposition as described in \citet{fan2019selective}. Let $\mathcal{F}$ be the function space expressible by a particular DNN model and $\widehat{Y}_{N}(\mathbf{s}_0)$ be the final prediction from the model based on $N$ observed locations. The following decomposition of the total risk between the true value $Y(\mathbf{s}_0)$ and the prediction $\widehat{Y}_{N}(\mathbf{s}_0)$ implies three sources of errors:
\[
R\{Y(\mathbf{s}_0),\widehat{Y}_{N}(\mathbf{s}_0)\}=\underset{\mbox{approximation error}}{R\{\underbrace{Y(\mathbf{s}_0),\widehat{Y}_{\mathcal{F}}^{\mathrm{opt}}(\mathbf{s}_0)\}}}+\underset{\mbox{estimation error}}{\underbrace{R\{\widehat{Y}_{\mathcal{F}}^{\mathrm{opt}}(\mathbf{s}_0),\widehat{Y}^{\mathrm{opt}}_{N}(\mathbf{s}_0)\}}}+\underset{\mbox{optimization error}}{\underbrace{R\{\widehat{Y}^{\mathrm{opt}}_{N}(\mathbf{s}_0),\widehat{Y}_{N}(\mathbf{s}_0)\}}}.
\] 
The approximation error relates to the model capacity and is defined as the risk between the true process $Y(\mathbf{s}_0)$ and the optimal predictor $\widehat{Y}_{\mathcal{F}}^{\mathrm{opt}}(\mathbf{s}_0) = \underset{\widehat{Y}(\mathbf{s}_0)\in \mathcal{F} }{\mathrm{argmin}} ~ {R}(\widehat{Y}(\mathbf{s}_0),Y(\mathbf{s}_0))$ as a function in $\mathcal{F}$. The estimation error is defined as the risk between $\widehat{Y}^{\mathrm{opt}}_{N}(\mathbf{s}_0)$ and $\widehat{Y}_{\mathcal{F}}^{\mathrm{opt}}(\mathbf{s}_0)$, where $\widehat{Y}^{\mathrm{opt}}_{N}(\mathbf{s}_0)=\widehat{Y}_{N}(\mathbf{s}_0;\hat{\boldsymbol{\theta}})$, with $\hat{\boldsymbol{\theta}}=\underset{\boldsymbol{\theta}}{\mathrm{argmin}} \frac{1}{N}\sum_{n=1}^N L\{\widehat{Y}_{N}(\mathbf{s}_n;\boldsymbol{\theta}),z(\mathbf{s}_n)\}$; this type of error is affected by the complexity of $\mathcal{F}$ and relates to the generalization power of the model. The optimization error is the empirical risk between $\widehat{Y}^{\mathrm{opt}}_{N}(\mathbf{s}_0)$ and $\widehat{Y}_{N}(\mathbf{s}_0)$.

The function class of Kriging prediction in Equation \eqref{eq:Kriging}, $\mathcal{F}_{\rm{UK}}$, can be viewed as the space of linear functions of $\mathbf{x}(\mathbf{s}_0)$ and $\mathbf{z}$ taking the form $ \mathbf{x}(\mathbf{s}_0)^T\boldsymbol{\beta}+ \mathbf{z}^T\boldsymbol{\gamma}$, while the function class of DeepKriging, $\mathcal{F}_{\rm{DK}}$, is the function space generated by the DNN described in \eqref{eq:DK}. The universal approximation theorem (Theorem 2.3.1 of \citet{csaji2001approximation}) claims that every continuous function of the features $\mathbf{x}_{\phi}(\mathbf{s})$, denoted as $\mathbb{C}(\mathbf{x}_{\phi})$, can be arbitrarily well approximated with a feed-forward neural network with a single hidden layer that contains finite number of hidden neurons and with arbitrary activation function. This indicates that the optimal DeepKriging prediction with a single hidden layer and finite loss function has the largest model capacity in $\mathbb{C}(\mathbf{x}_{\phi})$, that is, $\widehat{Y}_{\mathcal{F}_{\rm{DK}}}^{\mathrm{opt}}(\mathbf{s}_0)=\widehat{Y}_{\mathbb{C}(\mathbf{x}_{\phi})}^{\mathrm{opt}}(\mathbf{s}_0)$. This result holds for any type of data (i.e., continuous or discrete) and for any type of task (i.e., regression or classification). Therefore, the optimal DeepKriging prediction has larger capacity than the Kriging prediction in terms of minimizing the approximation error, i.e., $\mathbb{E}\{L(\widehat{Y}_{\mathcal{F}_{\rm{DK}}}^{\mathrm{opt}}(\mathbf{s}_0),Y(\mathbf{s}_0))\}\leq \mathbb{E}\{L(\widehat{Y}_{\mathcal{F}_{\rm{UK}}}^{\mathrm{opt}}(\mathbf{s}_0),Y(\mathbf{s}_0))\}$. The detailed proof is provided in Section~S1.2 in the Supplementary Materials. Similarly, the optimal DeepKriging prediction also has larger model capacity than the FRK prediction. FRK can be seen as DeepKriging with a single hidden layer containing finite number of neurons and a linear activation function. By allowing for a large number of basis functions, multiple layers, more flexible activation functions and a wide network, DeepKriging yields non-linear predictions that can appropriately capture the spatial dependence in the spatial process.

\subsection{DeepKriging as a Gaussian Process}\label{sec:GP}
\citet{neal1994priors} showed that a single-layer fully-connected neural network with an i.i.d. prior over its parameters (i.e., weights and biases) is equivalent to a Gaussian process (GP), in the limit of infinite network width (i.e., infinite number of hidden neurons). Later, \citet{lee2017deep} derived the exact equivalence between infinitely wide deep networks and GPs. Consequently, a similar correspondence to GPs also holds for our DeepKriging network.

We start from a regression-type DeepKriging model with a singe hidden layer containing $N_1$ neurons. The input features are $\mathbf{x}_{\phi}(\mathbf{s})=(\mathbf{x}(\mathbf{s})^T, \boldsymbol{\phi}(\mathbf{s})^T)^T\in\mathbb{R}^{P+K}$, and the output is $\hat{Y}_{\rm{DK}}(\mathbf{s})=b^1 + \sum_{j=1}^{N_1}w^{1}_{j} a^{1}_{j}(\mathbf{s})$, where 
$a^{1}_{j}(\mathbf{s})=\psi_1(b^{0}_{j} + \sum_{i=1}^{P+K}w^{0}_{ji}\mathbf{x}^{(i)}_{\phi}(\mathbf{s}))$, with $\mathbf{x}^{(i)}_{\phi}(\mathbf{s})$ being the $i$-th component of $\mathbf{x}_{\phi}(\mathbf{s})$. Weights ($w^{1}_{j}$, $w^{0}_{ji}$) and biases ($b^1$, $b^{0}_{j}$) are independent and randomly drawn to have zero mean and variances $\sigma^{2}_{w}/N_1$ and $\sigma^{2}_{b}$, respectively. Consequently, the post-activations $a^{1}_{j}$ and $a^{1}_{j'}$ are independent for $j\neq j'$. Moreover, since $\hat{Y}_{\rm{DK}}(\mathbf{s})$ is a sum of i.i.d terms, it follows from the Central Limit Theorem that in the limit of infinite width $N_{1}\to\infty$, $\hat{Y}_{\rm{DK}}(\mathbf{s})$ will be Gaussian distributed. Likewise, from the multi-dimensional Central Limit Theorem, any finite collection of $\{\hat{Y}_{\rm{DK}}(\mathbf{s}_1),\hat{Y}_{\rm{DK}}(\mathbf{s}_2),\cdots,\hat{Y}_{\rm{DK}}(\mathbf{s}_n)\}$ will have a joint multivariate Gaussian distribution, which is exactly the definition of a GP. Therefore, we conclude that with sufficiently large $N_1$, $\hat{Y}_{\rm{DK}}$ is a GP with zero mean and covariance function 
\begin{equation*}
C^{1}(\mathbf{s},\mathbf{s'})=E\{\hat{Y}_{\rm{DK}}(\mathbf{s})\hat{Y}_{\rm{DK}}(\mathbf{s'})\}=\sigma^{2}_{b}+\sigma^{2}_{w}E\{a^{1}_{j}(\mathbf{s})a^{1}_{j}(\mathbf{s'})\}=\sigma^{2}_{b}+\sigma^{2}_{w}C(\mathbf{s},\mathbf{s'}),
\label{eq:covariance} 
\end{equation*}
where $C(\mathbf{s},\mathbf{s'})$ is obtained by integrating against the distribution of $w^{0},\;b^0$ as in \citet{neal1994priors}.

For DeepKriging with deeper layers,  the induced covariance function can be obtained in a recursive way according to \citet{lee2017deep}: 
\begin{equation}
C^l(\mathbf{s},\mathbf{s'})=\sigma_b^2 + \sigma_w^2 F_{\psi}(C^{l-1}(\mathbf{s},\mathbf{s'}),C^{l-1}(\mathbf{s},\mathbf{s}),C^{l-1}(\mathbf{s'},\mathbf{s'})),
\label{eq:Cl}
\end{equation}
where $F_{\psi}(\cdot)$ is a deterministic function that depends only on the activation function ${\psi}$. An iterative series of computations can be performed to obtain the covariance $C^{L}$ for the GP describing the network’s final output, $\hat{Y}_{\rm{DK}}(\mathbf{s})$. For the base case, $C^0(\mathbf{s},\mathbf{s'})=\sigma_b^2 + \sigma_w^2\{\mathbf{x}_\phi(\mathbf{s})^{T}\mathbf{x}_\phi(\mathbf{s'})/(P+K)\}$. The aforementioned results require the assumption of infinitely many hidden neurons in each layer. However, when the prior distribution of weights and biases is Gaussian, this condition is not needed.

For certain activation functions, Equation \eqref{eq:Cl} can be computed analytically. The simplest case occurs when the activation function is an identity function $\psi_l(x)=x$ and no covariates effect exists. Then $\hat{Y}_{\rm{DK}}(\mathbf{s})$ is a linear function of the basis functions $\boldsymbol{\phi}(\mathbf{s})$, i.e., $\hat{Y}_{\rm{DK}}(\mathbf{s})= b + \mathbf{w}^T\boldsymbol{\phi}(\mathbf{s})$, where $b$ and $\mathbf{w}$ are combined biases and weights, respectively. In this case, the induced covariance function of $\hat{Y}_{\rm{DK}}$ is given by $C^{L}(\mathbf{s},\mathbf{s'})=\sigma_{b}^2+\sigma_{w}^2\boldsymbol{\phi}(\mathbf{s})^T\boldsymbol{\phi}(\mathbf{s'})$,
which is the basis approximation of a spatial covariance function. 

In the case of ReLU non-linearity, Equation \eqref{eq:Cl} has a closed form of the well-known arc-cosine kernel \citep{cho2009kernel}:
\begin{equation*}
C^l(\mathbf{s},\mathbf{s'})=\sigma_b^2 + \frac{\sigma_w^2}{2\pi}\sqrt{C^{l-1}(\mathbf{s},\mathbf{s})C^{l-1}(\mathbf{s'},\mathbf{s'})}\left\{\sin \left(\theta^{l-1}_{\mathbf{s},\mathbf{s'}}\right) +(\pi-\theta^{l-1}_{\mathbf{s},\mathbf{s'}})\cos \left(\theta^{l-1}_{\mathbf{s},\mathbf{s'}}\right)\right\},
\end{equation*}
where $\theta^{l}_{\mathbf{s},\mathbf{s'}}=\cos^{-1}(C^l(\mathbf{s},\mathbf{s'})/\sqrt{C^l(\mathbf{s},\mathbf{s})C^l(\mathbf{s'},\mathbf{s'})})$.
When no analytic form of the resulted covariance function exists, it can be computed numerically, as described in \citet{lee2017deep}.

Consider a regression-type DeepKriging model with a single hidden layer and no covariates effects. It can be shown that with infinitely many hidden neurons, the covariance function of the output $\hat{Y}_{\rm{DK}}(\mathbf{s})$ for any two nearby locations has the form 
\begin{equation}
C(\mathbf{s},\mathbf{s'}) = v(\mathbf{s})+v(\mathbf{s'})-c\|\boldsymbol{\phi}(\mathbf{s})-\boldsymbol{\phi}(\mathbf{s'})\|^2,
\label{eq:cov}
\end{equation}
where $\boldsymbol{\phi}(\mathbf{s})$ is the basis vector at location $\mathbf{s}$, $v(\mathbf{s})>0$ is related to the variance when $\mathbf{s} = \mathbf{s'}$, and $c$ is the scaling parameter. The proof is provided in Section~S1.3 in the Supplementary Materials. As a special case, if only the coordinates are used in the features, then $\|\boldsymbol{\phi}(\mathbf{s})-\boldsymbol{\phi}(\mathbf{s'})\|^2= \|\mathbf{s}-\mathbf{s'}\|^2$, $v(\mathbf{s})=v(\mathbf{s'})=v$ and thus $C(\mathbf{s},\mathbf{s'}) = v-c \|\mathbf{s}-\mathbf{s'}\|^2$, which contains less information than in Equation \eqref{eq:cov}. Therefore, the embedding layer in DeepKriging brings more flexible spatial covariance structures than simply using the coordinates.

Further, we can show how the DeepKriging induced covariance function can approximate the common stationary covariance functions in spatial statistics. Let the basis functions be $\phi_l(\mathbf{s}) = k(\mathbf{s},\mathbf{u}_l)$ based on a certain kernel function $k(\cdot,\cdot)$ and knot $\mathbf{u}_l$, $l=1,\cdots,K$.  If the $\mathbf{u}_l$'s form a fine grid of knots covering the spatial domain, then
\begin{align*}
&\|\boldsymbol{\phi}(\mathbf{s})-\boldsymbol{\phi}(\mathbf{s'})\|^2 = \sum_{l=1}^K\{k(\mathbf{s},\mathbf{u}_l)-k(\mathbf{s'},\mathbf{u}_l)\}^2 \approx \int\{k(\mathbf{s},\mathbf{u})-k(\mathbf{s'},\mathbf{u})\}^2\mbox{d}\mathbf{u}\\
&=\int k(\mathbf{s},\mathbf{u})^2 + k(\mathbf{s'},\mathbf{u})^2-2k(\mathbf{s},\mathbf{u})k(\mathbf{s'},\mathbf{u})\mbox{d}\mathbf{u}.
\end{align*}

Note that the last term is the kernel convolution approximation to a covariance function. \cite{higdon2002space} shows that by selecting an appropriate kernel function, we can approximate any stationary covariance function based on the kernel convolution. Further, the induced covariance function of DeepKriging also possesses favorably physical interpretations. For example, DeepKriging can yield the Mat{\'e}rn covariance function, also commonly used in Kriging since it is related to a stochastic partial differential equation (SPDE) of Laplace type \citep{whittle1954stationary}. In addition, DeepKriging can induce a GP that approximates a fractional Brownian motion based on the example of DNN provided in \citet{neal2012bayesian}. 

\section{Simulation Studies}\label{sec:sims}
\subsection{DeepKriging on a 1-D Gaussian process}
We first consider the performance of DeepKriging when data are simulated from a 1-D stationary GP, where the Kriging prediction is optimal. We also compare DeepKriging to two naive DNNs: a DNN with only the intercept $x(s)=1$ as the input and a DNN with $x(s)=1$ and coordinate $s$ as the input. We also consider Kriging prediction with the true covariance function and that with an estimated Mat{\'e}rn covariance function.  
The simulation design is illustrated in Section S3.1 of the Supplementary Materials.  

Figure S1 in the Supplementary Materials shows the prediction for one of the sample datasets using each of the five prediction methods. The DNN with the intercept only predicts the mean of the process. Although including the coordinate $s$ in the DNN improves the prediction, it fails to capture the high-frequency variability and cannot reflect the spatial correlations of the true process. Moreover, DeepKriging prediction and the optimal Kriging prediction are almost overlapped.

To further validate the performance, we calculate the root MSE (RMSE) and mean absolute percentage error (MAPE) on the testing data over the 100 replicated samples in Table S1 in the Supplementary Material, where MAPE is defined as $\frac{1}{N_{\mathrm{test}}}\sum_{n=1}^{N_{\mathrm{test}}}\frac{Y_{n}^{\rm{pred}}-Y_{n}^{\mathrm{true}}}{Y_{n}^{\mathrm{true}}}$, $N_{\mathrm{test}}$ is the number of testing samples, $Y_{n}^{\mathrm{pred}}$ is the predicted value and $Y_{n}^{\mathrm{true}}$ is the true value. As the minimum-MSE predictor, the Kriging prediction with the true covariance function has the smallest RMSE as expected. The performance of DeepKriging is comparable to the two Kriging predictions and significantly outperforms the two naive DNN models. 
We also provide the results on the training set in Table S1 in the Supplementary Material. Again, the Kriging prediction with the true covariance function performs the best. The DeepKriging prediction is comparable to the optimal Kriging prediction, and it outperforms the Kriging prediction with an estimated covariance function and the two naive DNN models in terms of both RMSE and MAPE.

\subsection{DeepKriging on 2-D non-stationary data}
In this section, we evaluate the performance of DeepKriging on 2-D non-stationary data so that the procedure is designed to resemble the real data application in Section \ref{sec:application}. The simulation details are included in Section S3.2 of the Supplementary Material. 

We use the 10-fold cross-validation method to show the performance of DeepKriging, Kriging with an estimated stationary covariance function and the baseline DNN with only coordinates $s$ in the features. We calculate the RMSEs and MAPEs on the testing dataset, and show the results in Figure S2(b) and Table S2. We can see that in terms of RMSE, DeepKriging significantly outperforms Kriging in terms of RMSEs and MAPEs, since Kriging assumes a stationary covariance function while DeepKriging captures the non-stationarity in the data. In addition, the baseline DNN is better than Kriging in this example because the data are non-Gaussian and Kriging is no longer optimal. 
 Moreover, the baseline DNN performs worse than DeepKriging as expected. The MAPE from DeepKriging is lower than the baseline DNN but higher than Kriging; this can happen since we are using MSE as the loss function in DeepKriging so it not necessarily possesses the lowest MAPE. We also calculate the RMSEs and MAPEs on the training dataset (see Table S2). Kriging outperforms the other two models in terms of both metrics. This is because the errors for the training dataset can be viewed as the variance estimates of the assumed model, similar as in a regression model. Kriging tends to underestimate such a variance, leading to a worse prediction on the testing dataset.

Additional simulations (see Section~S3 of the Supplementary Material) are conducted to show that DeepKriging is non-linear in observation whereas Kriging is linear. Furthermore, the comparison of computation time based on the same simulation study shows that Kriging is faster for small sample sizes ($N<1,500$), but DeepKriging is much more scalable when the sample size increases. This is because when the sample size is small, the computation time is still under control for Kriging, but for DeepKriging, the number of parameters is large due to the large width and depth of the network, making the computation time longer than Kriging. When the sample size increases, the computational burden of both methods also increases, but for DeepKriging, we can use parallel computing for the data with CPUs or GPUs to largely accelerate the computation. Therefore, our DeepKriging method is much more scalable to large data sizes. For example, when $N=12,800$, it takes more than 1.5 hours ($5,663$ seconds) to implement a Kriging model, while DeepKriging only takes 3.5 minutes ($214$ seconds) without GPU acceleration and 1.5 minutes ($94$ seconds) with a Tesla P100 GPU. 

\section{Application}\label{sec:application}
\subsection{Challenges of predicting PM$_{2.5}$ concentration}
PM$_{2.5}$, fine particulate matter of less than $2.5 \,\mu{m}$, is a harmful air pollutant. Its adverse effects are associated with many diseases such as respiratory disease \citep{peng2009emergency} and myocardial infarction \citep{peters2001increased}; see the review by \citet{world2013health}. Therefore, it is essential to obtain a high-resolution map of PM$_{2.5}$ exposure in order to assess its impact.
The measurements from monitoring networks are the best characterization of PM$_{2.5}$ concentration at a given time and location. However, data from monitoring locations are often sparsely distributed so that they are out of spatial and temporal alignment with health outcomes. Meanwhile, it is known that PM$_{2.5}$ concentration is associated with meteorological conditions such as temperature and relative humidity \citep{jacob2009effect}, where the meteorological data or data products are often easy to access with good spatial coverage and resolutions. Hence, the interpolation of PM$_{2.5}$ concentration by making use of data from monitoring networks and other meteorological data has been a promising field of research \citep{di2016assessing}, where spatial prediction plays a central role.

The modeling and prediction of PM$_{2.5}$ concentration are challenging. First, PM$_{2.5}$ concentration data are obviously non-Gaussian, and thus classical Kriging methods are inappropriate here. Second, PM$_{2.5}$ data from monitoring stations are irregular and sparse, but many interpolation methods require lattice data. Third, it is more important but challenging to understand the risk of high pollution and predict pollution levels, such as being low, medium and high; statistically, these two questions are related to estimating the probability over a threshold and a classification problem, respectively. Quantile regression and convolutional neural networks have been employed to overcome some of the above issues \citep{reich2011bayesian, porter2015investigating, di2016assessing}. However, a unified method to handle all of the aforementioned tasks has not yet been sufficiently developed.

\subsection{Data and preprocessing}
To tackle the above-mentioned problems, we apply the proposed DeepKriging method to the spatial prediction of PM$_{2.5}$ concentrations based on meteorological variables. Meteorological data are obtained from the NCEP North American Regional Reanalysis (NARR) product. Reanalysis is a gridded dataset that represents the state of the atmosphere, incorporating observations and outputs of numerical weather prediction models from past to present-day. Reanalysis data are often used to represent the ``true state'' of the atmosphere according to observations, and thus we use the reanalysis data as the ``observed data'' for the covariates. A total of six meteorological variables are used in this study: 1) air temperature at $2$ m, 2) relative humidity at $2$ m, 3) accumulated total precipitation, 4) surface pressure, 5) u-component of wind, and 6) v-component of wind at $10$ m. The covariates from the NARR product are gridded data on June 05, 2019 with a spatial resolution of about $32\times 32$ km covering the continental U.S., containing $7,706$ gridded cells in total. Since the units of the meteorological variables are different, we use min-max normalization to re-scale the data before implementing the models. Daily averaged data of PM$_{2.5}$ concentrations are observed from $841$ monitoring stations. Since the coordinates from NARR and those from stations are not identical and some of stations are too close to each other, we keep the spatial resolution of NARR and average the PM$_{2.5}$ measurements of nearby monitoring stations in the same grid cell. After the matching, 604 grid cells remain for the model training, with the PM$_{2.5}$ concentration value at each location shown in Figure~\ref{fig:app}(a). Our goal is to predict the PM$_{2.5}$ concentrations at any $s_0$ of the other $7,706-604=7,102$ locations where the PM$_{2.5}$ concentrations are not observed but the covariates are provided by the reanalysis data. 

\subsection{Model fitting and results}
Our aim is to predict the PM$_{2.5}$ concentration values at unobserved grid cells where the six meteorological variables are provided. We use the 10-fold cross-validation to verify the performance of DeepKriging. For comparison purposes, we also show the results from Kriging and the baseline DNN with the six covariates and coordinates. We calculate the MSEs and MAEs as the validation criterion, shown in the first two rows of Table \ref{tab:app}, which imply that DeepKriging outperforms the baseline DNN and Kriging.

To assess the risk of high PM$_{2.5}$ pollution, we can use DeepKriging for spatial data classification. Specifically, we threshold the PM$_{2.5}$ concentrations by $12.0$~$\mu g/m^3$, which is the threshold between ``good'' and ``moderate'' levels for the daily mean of EPA national ambient air quality standards (NAAQS) \citep{standard}. Based on the classified data, we can implement a binary classification with DeepKriging by assuming the actual values of PM$_{2.5}$ concentration to be unknown. A direct comparison to Kriging is not feasible since Kriging is not suitable for binary classification. Instead, we predict the continuous PM$_{2.5}$ concentrations using Kriging and then classify the predictions by thresholding them at $12.0$~$\mu g/m^3$. We then use the 10-fold cross-validation to show the classification accuracy, presented in the last row of Table \ref{tab:app}. We can see that DeepKriging significantly outperforms Kriging and baseline DNN in terms of the classification accuracy.
 \def\arraystretch{1.2}
  \begin{table}[h]
    \caption{\baselineskip=18pt \label{tab:app}Model performance based on the 10-fold cross-validation. MSEs and MAEs of the predictions, as well as classification accuracy (ACC) for predicting PM$_{2.5}$ concentrations above $12.0\;\mu g/m^3$ are used as the validation criteria. Mean and standard deviation (SD) of the 10 sets of validation errors or accuracy are provided in the table.}
    \vspace*{.5cm}
  \centering
\begin{tabular}{c|c| c| c| c |c | c}
\hline
   Parameters &\multicolumn{2}{c|}{DeepKriging} & \multicolumn{2}{c|}{Baseline DNN} &\multicolumn{2}{c}{Kriging}   \\&Mean & SD &  Mean & SD&Mean & SD\\
   \hline
   MSE & $\textbf{1.632}$ &  $.572$ &  $3.632$ &  $.925$ &3.361&.773  \\
   MAE &  $\textbf{.892}$ &  $.103$ &  $1.448$ &   $.162$&1.365&.178    \\
   ACC &  $\textbf{95.2\%}$ &  $2.6\%$ &  $89.6\%$ &   $4.8\%$&88.5\%&4.6\%    \\
   \hline
\end{tabular}
\label{tab:app}
\end{table}

\begin{figure}[h]
\centering
\includegraphics[width=\textwidth]{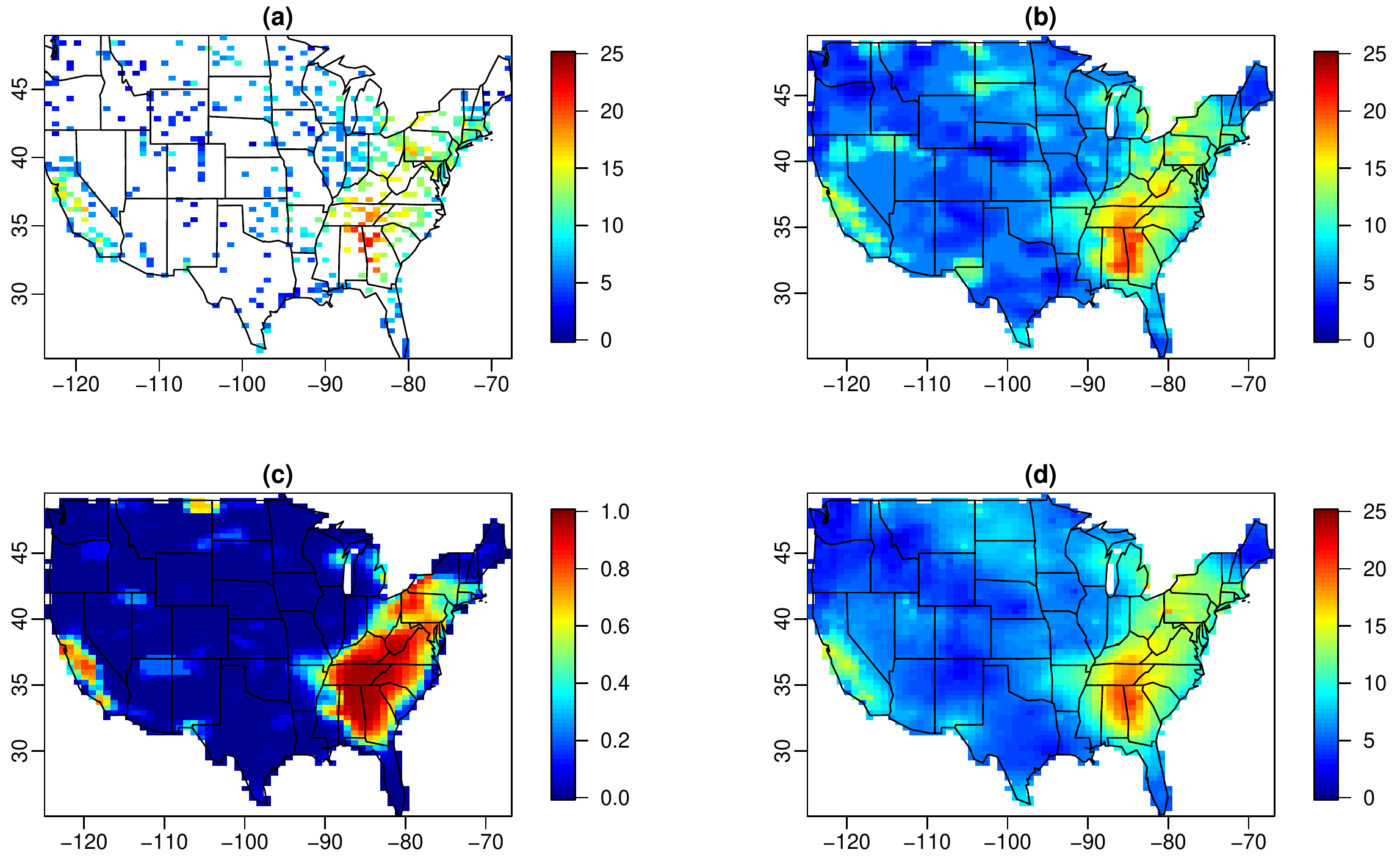}
\vspace*{-1cm}
\caption{\baselineskip=20pt (a) PM$_{2.5}$ concentration ($\mu g/m^3$) collected from monitoring stations. (b) Predicted PM$_{2.5}$ concentration using DeepKriging.  (c) Predicted risk of high pollution $\mathbb{P}\{\mathrm{PM}_{2.5}>12\;\mu g/m^3\}$ based on distribution prediction using DeepKriging. (d) Predicted PM$_{2.5}$ concentration using Kriging.}
\label{fig:app}
\end{figure}

Based on the model fitting, we can further predict the value of PM$_{2.5}$ concentration, the level of pollution and the risk of high pollution level over the threshold $12$ $\mu g/m^3$ at unobserved locations based on the NARR data. Figure~\ref{fig:app}(a) shows the raw PM$_{2.5}$ station data from the AQS database. Figure~\ref{fig:app}(b) shows a smooth map of the predicted PM$_{2.5}$ concentration from DeepKriging. We also provide the distribution prediction (details and algorithms are included in Section~S5 in the Supplementary Material) in order to obtain the predicted risk defined as $\mathbb{P}\{\mathrm{PM}_{2.5}>12\;\mu g/m^3\}$, shown in Figure~\ref{fig:app}(c). This map implies that high PM$_{2.5}$ pollution risks exist over a vast area of Eastern US. We further compare the results to the Kriging prediction in Figure~\ref{fig:app}(d), which implies that DeepKriging provides more local features/patterns than Kriging.

\section{Discussion}\label{sec:discussion}
In this work, we have proposed a new spatial prediction model using deep neural networks which incorporates the spatial dependence by a set of basis functions. Our method does not assume parametric forms of covariance functions or data distributions, and is generally compatible with non-stationarity, non-linear relationships, and non-Gaussian data. Uncertainty quantification can be provided based on our DeepKriging framework using the distribution prediction method detailed in Section~S5 in the Supplementary Materials.

Classical Kriging methods consider their predictions as linear combinations of observations, which impedes their interaction with several machine learning frameworks. Some evidence of the equivalence between Kriging and radial basis functions interpolation has been known since 1981 in \citet{matheron1981splines}. However, without the modern machine learning tools, only a linear combination and a limited number of radial basis functions have been investigated, which are viewed as a less favorable choice to Kriging \citep{dubrule1983two, dubrule1984comparing}. This work has provided a new perspective on deep learning in spatial prediction with a large number of basis functions. We have shown that the proposed method is superior to Kriging in many aspects both theoretically and numerically in our simulation and real application. For instance, DeepKriging is more scalable for large datasets and suits for more data types than Kriging. DeepKriging also has a GP representation with flexible spatial covariance structures, which enables Bayesian inference on regression tasks by evaluating the corresponding GP. More importantly, the proposed DeepKriging framework connects the regression-based prediction and spatial prediction so that many other machine learning algorithms can be applied.

In general applications, it is possible that the covariates at the new location $\mathbf{s}_0$ are not observed. One promising approach for coping with this problem is to find the true values of the missing covariates for a subset of the observations and then train a machine learning algorithm to predict the values of those covariates for the rest (see, e.g., \cite{imai2016improving}). However, \cite{fong2021machine} showed that plugging in these predictions without regard for prediction error renders regression analyses biased, inconsistent, and overconfident. They described a procedure to avoid these inconsistencies. This approach combines a new sample splitting scheme and a general method of moments (GMM) estimator to make an efficient and consistent estimator. Overall, it is non-trivial to address the problem of missing covariates: intuitive strategies such as plugging in machine learning predictions lead to bias and inconsistency, while the implementation of a more complicated method such as that in \cite{fong2021machine} requires extra assumptions (e.g., the exclusion restriction condition) and increases computational burden. If the goal is to predict both the response and the covariates instead, a multivariate version of DeepKriging could be developed. These are left as our future work.

\section*{Supplementary Materials}
The Supplementary Material contains details referenced in the main manuscript, including the proofs of the lemmas and theorems (Section S1), the settings for the DeepKriging network structure (Section S2), details of the simulation studies (Section S3), additional simulation studies (Section S4), distribution prediction and uncertainty quantification (Section S5), and the source codes and data for reproducible research (Section S6).
\par
\section*{Acknowledgments}

This research is supported by the National Key Research and Development Program (2021YFA1000101), Zhejiang Provincial Natural Science Foundation of China (LZJWY22E090009), Natural Science Foundation of Shanghai (22ZR1420500), the Open Research Fund of Key Laboratory of Advanced Theory and Application in Statistics and Data Science-MOE, ECNU and King Abdullah University of Science and Technology (KAUST), Office of Sponsored Research (OSR) under Award No: OSR-2019-CRG7-3800.
\par


\bibhang=1.7pc
\bibsep=2pt
\fontsize{9}{14pt plus.8pt minus .6pt}\selectfont
\renewcommand\bibname{\large \bf References}
\expandafter\ifx\csname
natexlab\endcsname\relax\def\natexlab#1{#1}\fi
\expandafter\ifx\csname url\endcsname\relax
  \def\url#1{\texttt{#1}}\fi
\expandafter\ifx\csname urlprefix\endcsname\relax\def\urlprefix{URL}\fi




\newpage
\vskip .65cm
\noindent
Wanfang Chen: Academy of Statistics and Interdisciplinary Sciences, East China Normal University, Shanghai, China
\vskip 2pt
\noindent
E-mail: wfchen@fem.ecnu.edu.cn

\noindent
Yuxiao Li: Statistics Program, King Abdullah University of Science and Technology, Thuwal, Saudi Arabia
\vskip 2pt
\noindent
E-mail: yuxiao.li@kaust.edu.sa

\noindent
Brian J Reich: Department of Statistics, North Carolina State University, Raleigh, North Carolina, U.S.A.
\vskip 2pt
\noindent
E-mail: brian\_reich@ncsu.edu

\noindent
Ying Sun (corresponding author): Statistics Program, King Abdullah University of Science and Technology, Thuwal, Saudi Arabia
\vskip 2pt
\noindent
E-mail: ying.sun@kaust.edu.sa; \;Phone: +966 (0) 56 898-5402;\;Tax: +966 (0) 12 808-0644

\newpage
 \centerline{\bf\large Supplementary Material}
\vspace{.35cm}
\fontsize{12}{14pt plus.8pt minus .6pt}\selectfont
\noindent
This supplementary material contains the proofs (Section S1), the settings for the DeepKriging network structure (Section S2), details of the simulation studies (Section S3), additional simulation studies (Section S4), distribution prediction and uncertainty quantification (Section S5), and the source codes and data for reproducible research (Section S6).
\par

\setcounter{section}{0}
\setcounter{equation}{0}
\setcounter{figure}{0}
\setcounter{table}{0}
\def\theequation{S\arabic{section}.\arabic{equation}}
\def\thesection{S\arabic{section}}
\def\thefigure{S\arabic{figure}}
\def\thetable{S\arabic{table}}

\renewcommand{\theequation}{\thesection.\arabic{equation}}

\fontsize{12}{14pt plus.8pt minus .6pt}\selectfont

\baselineskip=24pt
\section{Proofs}\label{sec:appendix}
\subsection{Kriging prediction is linear in $\mathbf{x}_{\phi}(\mathbf{s})=(\mathbf{x}(\mathbf{s}),\boldsymbol{\phi}(\mathbf{s}))^{T}$}

\textbf{Proof}: Let the covariance matrix associated with the random process $\nu(\mathbf{s})$ be $\boldsymbol{\Sigma}^{\nu}$, where $\boldsymbol{\Sigma}^{\nu}_{i,j}=C(\mathbf{s}_i,\mathbf{s}_j) $. We can build a spatial random effect process $\mu(\mathbf{s})=\boldsymbol{\phi}(\mathbf{s})^T\boldsymbol{\eta}$ with $\mathrm{Cov}(\boldsymbol{\eta}) = \boldsymbol{\Phi}_R^{-1}\boldsymbol{\Sigma}^{\nu}(\boldsymbol{\Phi}_R^{-1})^{T}$, where $\boldsymbol{\Phi}=\{\boldsymbol{\phi}(\mathbf{s}_1),\dots,\boldsymbol{\phi}(\mathbf{s}_N)\}$ is an $N\times K$ basis matrix with rank $N$, and $\boldsymbol{\Phi}_R^{-1}$ is the right inverse of $\boldsymbol{\Phi}$. From the result of \citet{banerjee1973generalized}, the right inverse exists. Hence, $\mathrm{Cov}(\boldsymbol{\eta})$ is also a valid covariance matrix and $\mathrm{Cov}(\boldsymbol{\mu})=\mathrm{Cov}(\boldsymbol{\nu})= \boldsymbol{\Sigma}^{\nu}$, where $\boldsymbol{\mu}= \{{\mu}(\mathbf{s}_1),\dots,{\mu}(\mathbf{s}_N)\}^T$ and $\boldsymbol{\nu}= \{{\nu}(\mathbf{s}_1),\dots,{\nu}(\mathbf{s}_N)\}^T$.

Therefore, the Kriging prediction of $\mu(\mathbf{s})$ is the same as that of $\nu(\mathbf{s})$, i.e.,  
$\widehat{Y}_{\mu}^{\mathrm{UK}}(\mathbf{s}_0) = \widehat{Y}_{\nu}^{\mathrm{UK}}(\mathbf{s}_0)$. On the other hand, based on the spatial random effect representation of $\mu(\mathbf{s})$, we can get the equivalent fixed rank Kriging prediction shown in the Equation (5) in the manuscript, 
such that $\widehat{Y}_{\mu}^{\mathrm{UK}}(\mathbf{s}_0) = \widehat{Y}_{\mu}^{\mathrm{FRK}}(\mathbf{s}_0)=\mathbf{x}(\mathbf{s}_0)^T\widehat{\boldsymbol{\beta}} + \boldsymbol{\phi}(\mathbf{s}_0)^T\widehat{\boldsymbol{\alpha}}.\qed$

\subsection{Model capacity of DeepKriging compared to Kriging}
\textbf{Proof}: Without loss of generality, we consider the mean squared loss in the prediction. The universal approximation theorem (Theorem 2.3.1 of \citet{csaji2001approximation}) shows that $\forall f\in \mathbb{C}(\mathbf{x}_{\phi})$, $\forall \varepsilon>0$: $\exists n\in\mathbb{N}$ and certain choice of weights and biases, such that $|f(\mathbf{s}_0)-A_nf(\mathbf{s}_0)|<\varepsilon$, where $A_nf$ is the mapping from the standard multi-layer feed-forward networks with a single hidden layer that contains $n$ hidden neurons. Then we have 
\begin{equation}
	\displaystyle \lim_{n\to \infty} |f(\mathbf{s}_0) - A^{\mathrm{opt}}_nf(\mathbf{s}_0)|^2=0,
	\label{eq:universal}
\end{equation}
where $A^{\mathrm{opt}}_nf(\mathbf{s}_0)$ is the optimal neural network that approximating $f(\mathbf{s}_0)$.

In the context of spatial prediction, suppose the optimal prediction in $\mathbb{C}(\mathbf{x}_{\phi})$ is $\widehat{Y}_{\mathbb{C}(\mathbf{x}_{\phi})}^{\mathrm{opt}}(\mathbf{s}_0)$ that minimizes the mean squared loss, i.e.,
$\widehat{Y}_{\mathbb{C}(\mathbf{x}_{\phi})}^{\mathrm{opt}}(\mathbf{s}_0) =\underset{\widehat{Y}(\mathbf{s}_0)\in \mathbb{C}(\mathbf{x}_{\phi})}{\mathrm{argmin}} \mathbb{E}\{|\widehat{Y}(\mathbf{s}_0)-Y(\mathbf{s}_0)|^2\}$.
Then, the mean squared loss of $\widehat{Y}_{\mathcal{F}_{\rm{DK}}}^{\mathrm{opt}}(\mathbf{s}_0)$ satisfies
\[
0\leq \mathbb{E}\{|\widehat{Y}_{\mathcal{F}_{\rm{DK}}}^{\mathrm{opt}}(\mathbf{s}_0)-Y(\mathbf{s}_0)|^2\}-\mathbb{E}\{| \widehat{Y}_{\mathbb{C}(\mathbf{x}_{\phi})}^{\mathrm{opt}}(\mathbf{s}_0)-Y(\mathbf{s}_0)|^2\}
\leq \mathbb{E}\{|\widehat{Y}_{\mathcal{F}_{\rm{DK}}}^{\mathrm{opt}}(\mathbf{s}_0)-\widehat{Y}_{\mathbb{C}(\mathbf{x}_{\phi})}^{\mathrm{opt}}(\mathbf{s}_0)|^2\}.
\]
Let $f(\mathbf{s}_0) = \widehat{Y}_{\mathbb{C}(\mathbf{x}_{\phi})}^{\mathrm{opt}}(\mathbf{s}_0) \in \mathbb{C}(\mathbf{x}_{\phi})$ in \eqref{eq:universal}, then $A^{\mathrm{opt}}_nf(\mathbf{s}_0) =\widehat{Y}_{\mathcal{F}_{\rm{DK}}}^{\mathrm{opt}}(\mathbf{s}_0)$ based on the definition, hence $\displaystyle \lim_{n\to \infty} |\widehat{Y}_{\mathcal{F}_{\rm{DK}}}^{\mathrm{opt}}(\mathbf{s}_0)-\widehat{Y}_{\mathbb{C}(\mathbf{x}_{\phi})}^{\mathrm{opt}}(\mathbf{s}_0)|^2=0$. 
Also note that 
\begin{equation*}
	|\widehat{Y}_{\mathcal{F}_{\rm{DK}}}^{\mathrm{opt}}(\mathbf{s}_0)-\widehat{Y}_{\mathbb{C}(\mathbf{x}_{\phi})}^{\mathrm{opt}}(\mathbf{s}_0))|^2 
	\leq |\widehat{Y}_{\mathbb{C}(\mathbf{x}_{\phi})}^{\mathrm{opt}}(\mathbf{s}_0)-Y(\mathbf{s}_0)|^2 + |\widehat{Y}_{\mathcal{F}_{\rm{DK}}}^{\mathrm{opt}}(\mathbf{s}_0)-Y(\mathbf{s}_0)|^2
	\leq 2 |\widehat{Y}_{\mathcal{F}_{\rm{DK}}}^{\mathrm{opt}}(\mathbf{s}_0)-Y(\mathbf{s}_0)|^2,
\end{equation*}
and $\mathbb{E}\{ |\widehat{Y}_{\mathcal{F}_{\rm{DK}}}^{\mathrm{opt}}(\mathbf{s}_0)-Y(\mathbf{s}_0)|\}^2<\infty$ by assumptions. 
Using the dominated convergence theorem, we have $\displaystyle \lim_{n\to \infty} \mathbb{E}\{|\widehat{Y}_{\mathcal{F}_{\rm{DK}}}^{\mathrm{opt}}(\mathbf{s}_0)-\widehat{Y}_{\mathbb{C}(\mathbf{x}_{\phi})}^{\mathrm{opt}}(\mathbf{s}_0)|^2\}= \mathbb{E}\{\lim_{n\to \infty} |\widehat{Y}_{\mathcal{F}_{\rm{DK}}}^{\mathrm{opt}}(\mathbf{s}_0)-\widehat{Y}_{\mathbb{C}(\mathbf{x}_{\phi})}^{\mathrm{opt}}(\mathbf{s}_0)|^2\}=0$. 
So, $\widehat{Y}_{\mathcal{F}_{\rm{DK}}}^{\mathrm{opt}}(\mathbf{s}_0)=\widehat{Y}_{\mathbb{C}(\mathbf{x}_{\phi})}^{\mathrm{opt}}(\mathbf{s}_0)=\underset{Y(\mathbf{s})\in \mathbb{C}(\mathbf{x}_{\phi})}{\mathrm{argmin}} \mathbb{E}\{|\widehat{Y}(\mathbf{s}_0)-Y(\mathbf{s}_0)|^2\}\}.$

Since $\mathcal{F}_{\rm{UK}} \subset \mathbb{C}(\mathbf{x}_{\phi})$, we have 
\[
\mathbb{E}\{L(\widehat{Y}_{\mathcal{F}_{\rm{DK}}}^{\mathrm{opt}}(\mathbf{s}_0),Y(\mathbf{s}_0))\}= \mathbb{E}\{L(\widehat{Y}_{\mathbb{C}(\mathbf{x}_{\phi})}^{\mathrm{opt}}(\mathbf{s}_0),Y(\mathbf{s}_0))\}\leq \mathbb{E}\{L(\widehat{Y}_{\mathcal{F}_{\rm{UK}}}^{\mathrm{opt}}(\mathbf{s}_0),Y(\mathbf{s}_0))\}.\qed
\]

\subsection{Proof of Equation (3.7)}
\textbf{Proof}: Recall that the induced covariance function of $\hat{Y}_{\rm{DK}}(\mathbf{s})$ is $$C(\mathbf{s},\mathbf{s'})=\mathbb{E}\{{Y}(\mathbf{s}){Y}(\mathbf{s'})\}=\sigma_b^2 + \sigma_w^2 \mathbb{E}\{a_{j}^{1}(\mathbf{s})a_{j}^{1}(\mathbf{s'})\}.$$
According to \cite{neal1994priors},  $$\mathbb{E}\{a_{j}^{1}(\mathbf{s})a_{j}^{1}(\mathbf{s'})\}=\frac{1}{2}[\mathbb{V}\mathrm{ar}\{a_{j}^{1}(\mathbf{s})\}+\mathbb{V}\mathrm{ar}\{a_{j}^{1}(\mathbf{s'})\}]-\frac{1}{2}\mathbb{E}\{[a_{j}^{1}(\mathbf{s})-a_{j}^{1}(\mathbf{s'})]^2\}.$$
When $\mathbf{s}$ is close to $\mathbf{s'}$, we have $\psi_1(x)-\psi_1(y)= \alpha(x-y)$ for a smooth activation function, where $\alpha$ is a scaling coefficient.  When the covariate vector $\mathbf{x}(\mathbf{s})$ is not available, we have $a_{j}^{1}(\mathbf{s})=\psi_1(b^{0}_{j} +\sum_{k=1}^{K}w^{0}_{jk}\phi_{k}(\mathbf{s}))$. Thus we have
$$\mathbb{E}\{[a_{j}^{1}(\mathbf{s})-a_{j}^{1}(\mathbf{s'})]^2\}=(\alpha \sigma_w)^2 \sum_{k=1}^{K}\{\phi_{k}(\mathbf{s})-\phi_{k}(\mathbf{s'})\}^2.$$ 
Finally, the covariance function for nearby locations is
\begin{align*}
	C(\mathbf{s},\mathbf{s'})&=\sigma_b^2 + \frac{\sigma_w^2}{2} [\mathbb{V}\mathrm{ar}\{a_{j}^{1}(\mathbf{s})\}+\mathbb{V}\mathrm{ar}\{a_{j}^{1}(\mathbf{s'})\}]-(\alpha \sigma^2_w)^2 \sum_{k=1}^{K}\{\phi_{k}(\mathbf{s})-\phi_{k}(\mathbf{s'})\}^2\\ &\equiv v(\mathbf{s})+v(\mathbf{s'})-c\|\boldsymbol{\phi}(\mathbf{s})-\boldsymbol{\phi}(\mathbf{s'})\|^2.\hfill \qed
\end{align*}

\section{Settings for the DeepKriging network structure}\label{sec:setting}
The model settings for the DeepKriging network structure include the following considerations. First, when the sample size is small, we add a dropout layer to avoid overfitting. For a large dataset with Gaussian priors, dropout and other types of regularization are not very helpful. Second, batch-normalization is another useful strategy since the covariates typically have different units, and the scales of basis function may vary too much for irregularly spaced spatial data. Third, we normalize the covariates before we run the automated batch-normalization since the covariates and basis functions required different types of normalization. Last, when the data are irregularly spaced, the value of the compactly supported basis functions for some knots that are far apart from any other locations will be zero; in this case, we remove these knots and the relevant columns in the basis matrix. 

To summarize, the default setting of DeepKriging network is as follows:
\begin{enumerate}[(1)]
\itemsep-0.2em
\item Normalize (min-max normalization) the observed covariates $\mathbf{x}(\mathbf{s})$; 
\item  Build the embedding layer using a $3$ to $5$ level multi-resolution radial basis functions $\boldsymbol{\phi}(\mathbf{s})$ with the corresponding basis matrix $\boldsymbol{\Phi}$; the form of the basis function and the choice of the number of basis functions $K$ are illustrated at the beginning of Section~2.2; 
\item  Remove the all-zero columns of the basis matrix $\boldsymbol{\Phi}$; 
\item  Add the 1st dense layer with 100 hidden neurons and ReLu activation; 
\item  Add the 1st dropout layer with 0.5 dropout rate;  
\item  Add the 1st batch-normalization layer;
\item  Add the 2nd dense layer with 100 hidden neurons and ReLu activation;
\item  Add the 2nd dropout layer with 0.5 dropout rate; 
\item  Add the 3rd dense layer with 100 hidden neurons and ReLu activation; 
\item  Add the 2nd batch-normalization layer;
\item  Add the output layer. 
\end{enumerate}

The default number of epochs for DeepKriging is 200. For regression tasks, the loss function is set to be the mean squared error (MSE), while for classification tasks, the loss function is set to be the cross-entropy loss. For optimization, we use the Adam optimizer \citep{kingma2014adam}. One can tune the hyper-parameters in the network (e.g., the value of $K$,  the number of layers, the width of each layer, the dropout rate, the batch size and the number of epochs) in order for the optimization to converge and to achieve a better performance compared to other basic methods such as Kriging. 

\section{Details of Simulation Studies}
\subsection{DeepKriging on a 1-D Gaussian process}
We first consider the performance of DeepKriging when data are simulated from a 1-D stationary GP, where the Kriging prediction is optimal. We also compare DeepKriging (with $x(s)$ and an embedding layer with basis functions of coordinate $s$ as the input) to a DNN that does not account for the spatial dependence (with only $x(s)$ as the input), and a DNN that incorporates the spatial dependence by including directly the coordinates in the features (with $x(s)$ and coordinate $s$ as the input). Specifically, the data are generated from a GP with a constant mean:  $z({s})= \mu + \nu({s})+\varepsilon({s})$, where $\mu=1$, $\nu({s})$ is zero mean GP with an exponential covariance function $C({s},{s'})=\sigma^2\exp\{-|{s}-{s'}|/\rho\}$, where the variance $\sigma^2=1$ and the range parameter $\rho = 0.1$, and $\varepsilon({s})$ is a Gaussian white noise with the nugget variance $\tau^2=0.01$. We generate $100$ replicates for $\{z(s_1),\cdots,z(s_N)\}$ from the GP with $N=1,000$ equally spaced locations over $[0,1]$, with $800$ locations randomly selected as training data and the remaining treated as testing data. In this example, there are no observed covariates except for the intercept, i.e., $x(s)=1$ for any $s\in \mathbb{R}$. 
We also compare these models to Kriging prediction with the true exponential covariance function and that with an estimated Mat{\'e}rn covariance function where the smoothness parameter is set to $1.5$. The above predictions related to Gaussian processes and deep learning are implemented using GPy \citep{gpy2014} and Keras \citep{ketkar2017deep}, respectively. Since we have Gaussian priors, dropout and other types of regularization for DeepKriging are not needed. The number of hidden layers is set to be 7 in DeepKriging, which yields the best performance (in terms of RMSE) for both the training and testing datasets among networks with 1 to 10 hidden layers (results not shown). Each hidden layer contains $100$ neurons and uses the ReLu activation. The loss function is MSE, the number of epochs is set to be $100$ and the batch size is set to be $32$. A four-level multi-resolution model is used to generate the basis functions for DeepKriging, thus $K= 10 + 19 + 37 + 73 = 139$. 

\begin{figure}[!h] 
	\centering
	\includegraphics[width=0.75\linewidth]{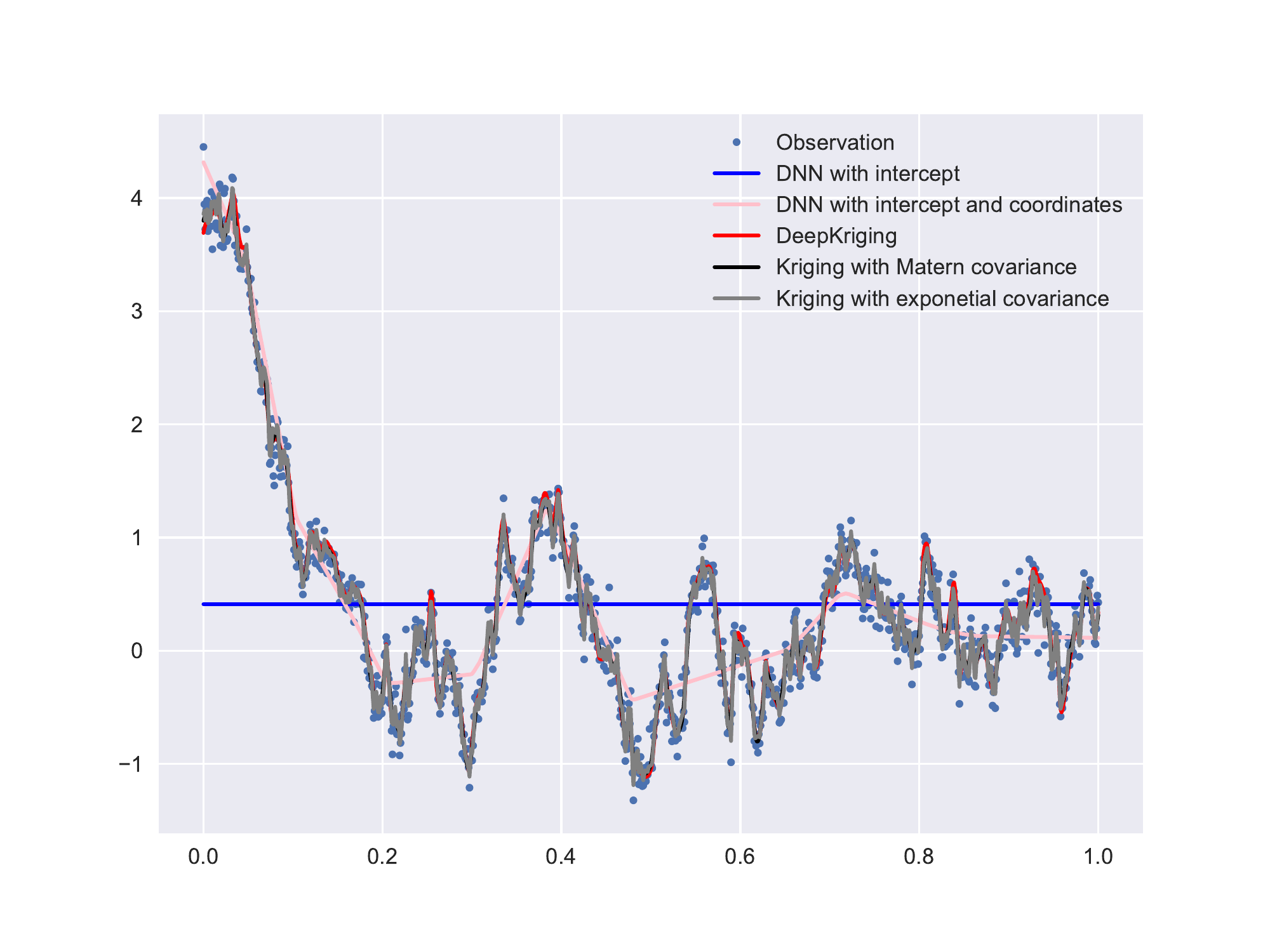}
	\vspace{-6mm}
	\caption{\baselineskip=20pt Prediction results for one simulated dataset generated from a GP. The blue dots are the simulated data (observations). The solid lines are the prediction curves from DNN with intercept only (blue line), DNN with intercept and coordinates (orange line), DeepKriging (red line), Kriging with the true exponential covariance function (grey line), and Kriging with an estimated Mat{\'e}rn covariance function (black line), respectively. }
	\label{fig:1dsims}
\end{figure}

Figure~\ref{fig:1dsims} shows the prediction for one of the sample datasets using each of the five prediction methods. Table~\ref{tab:1D} shows the root mean squared error (RMSE) and mean absolute percentage error (MAPE) on the both the training and testing sets over the 100 replicated samples. 

\begin{table}[!h]
	\caption{\baselineskip=16pt \label{tab:1D}Mean and standard deviations (SD) of the root mean squared errors (RMSEs) and mean absolute percentage errors (MAPEs) on both the training and testing sets across the 100 datasets from the five predictions of a Gaussian process. Kriging I and II are the Kriging prediction with the true exponential covariance function and that with an estimated Mat{\'e}rn covariance function, respectively. DeepKriging prediction is based on the MSE loss. DNN I and DNN II are the DNN prediction with intercept only and that with intercept and coordinates, respectively.}
	\vspace{2mm}
	\centering
	\renewcommand{\arraystretch}{1.2}
	\resizebox{\textwidth}{!}{%
		\begin{tabular}{c|c| c| c| c |c | c| c | c| c | c}
			\hline
			\multirow{2}{13mm}{Models} &\multicolumn{2}{c|}{Kriging I} & \multicolumn{2}{c|}{Kriging II} &\multicolumn{2}{c|}{DeepKriging} &\multicolumn{2}{c|}{DNN I} &\multicolumn{2}{c}{DNN II} \\
			&Mean & SD & Mean & SD & Mean & SD & Mean & SD & Mean & SD\\
			\hline
			\multicolumn{11}{l}{Training set}\\
			\hline
			RMSE & $\textbf{.068}$ &  $.002$ &  $.125$ &  $.006$ &.083&.009&.887&.183&.280&.029\\
			\hline
			MAPE &  $\textbf{.431}$ &  $1.639$ &  $.890$ &   $3.571$&.851&3.834&5.158&9.538&1.779&4.475\\
			\hline
			\multicolumn{11}{l}{Testing set}\\
			\hline
			RMSE & $\textbf{.158}$ &  $.008$ &  $.164$ &  $.008$ &.258&.015&.885&.185&.292&.031\\
			\hline
			MAPE &  $\textbf{.536}$ &  $.504$ &  $.570$ &   $.548$&.638&.540&3.123&2.755&.942&.843\\
			\hline
		\end{tabular}
	}
\end{table}

\subsection{DeepKriging on 2-D non-stationary data}
\begin{figure}[h!] 
	\centering
	\includegraphics[width = .46\linewidth]{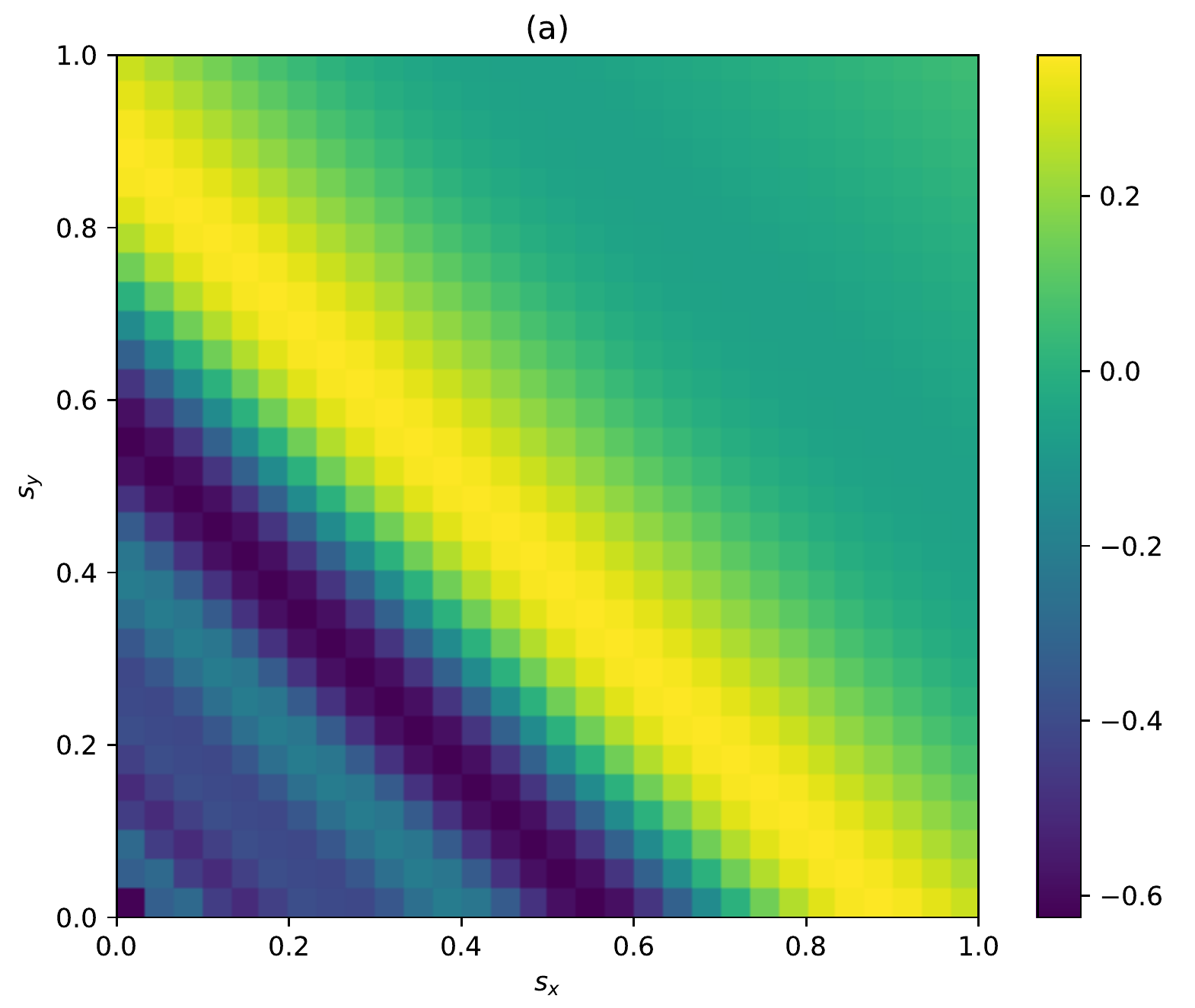}
	\includegraphics[width = .53\linewidth]{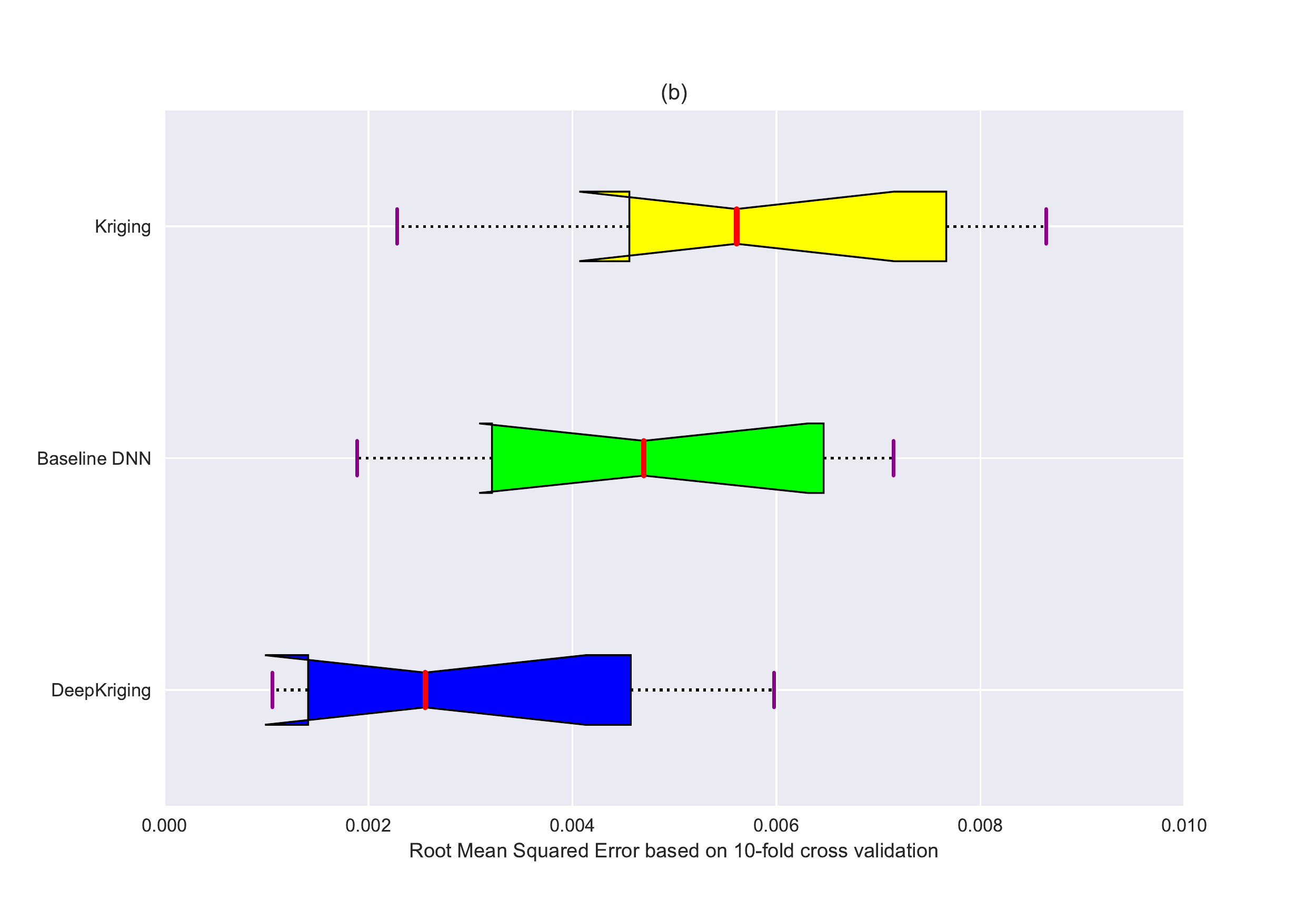}
	\vspace{-10mm}
	\caption{\baselineskip=20pt (a) Visualization of the simulated data generated from $Y(\mathbf{s}) = \sin\{30(\bar{s}-0.9)^4\}\cos\{2(\bar{s}-0.9)\}+(\bar{s}-0.9)/2$, where $\mathbf{s}=(s_x,s_y)^T\in [0,1]^2$ and $\bar{s}= (s_x+s_y)/2$. (b) Boxplots of the 10 RMSEs based on the 10-fold cross-validations from DeepKriging (blue), baseline DNN (green), and Kriging (yellow).}
	\label{fig:2dsims}
\end{figure}

In this section, we evaluate the performance of DeepKriging on 2-D non-stationary data so that the procedure is designed to resemble the real data application in Section 5 of the main manuscript. The goal is to predict values from the true process: $Y(\mathbf{s}) = \sin\{30(\bar{s}-0.9)^4\}\cos\{2(\bar{s}-0.9)\}+(\bar{s}-0.9)/2$, where $\mathbf{s}=(s_x,s_y)^T \in \mathbb{R}^2$ and $\bar{s}= (s_x+s_y)/2$. A similar example is evaluated in many computer experiments \citep{ba2012composite}, where both Kriging and neural networks are popularly applied. In our simulation, $N=900$ observations are sampled on a 30$\times$ 30 square grid of locations spanning $[0,1]^2$; see Figure~\ref{fig:2dsims}(a). This process presents obvious spatial non-stationarity; for example, the smoothness over the region $[0,0.4]^2$ is significantly smaller than that over $[0.4,1]^2$. 

The network structures of DeepKriging as well as the baseline DNN are the default setting as stated in Section~\ref{sec:setting}, except that the batch size is set to be 64, and the number of hidden layers is set to be 4 in DeepKriging, which yields the best performance (in terms of RMSE) for both the training and testing datasets among networks with 1 to 6 hidden layers (results not shown). A three-level multi-resolution model is used to generate the basis functions for DeepKriging, thus $K= 10^2 + 19^2 + 37^2 = 1,830$. We use the 10-fold cross-validation method to show the performance of DeepKriging, Kriging with an estimated stationary covariance function and the baseline DNN with only coordinates $s$ in the features. The boxplot of RMSEs are shown in Figure~\ref{fig:2dsims}(b). The RMSEs and MAPEs on both the training and testing sets are shown in Table~\ref{tab:2dsim}.

\begin{table}[h!]
	\caption{\baselineskip=16pt Model performance for both the training and testing sets based on 10-fold cross-validation. Data are generated from $Y(\mathbf{s}) = \sin\{30(\bar{s}-0.9)^4\}\cos\{2(\bar{s}-0.9)\}+(\bar{s}-0.9)/2$, where $\mathbf{s}=(s_x,s_y)^T\in [0,1]^2$ and $\bar{s}= (s_x+s_y)/2$. RMSEs and MAPEs are computed, and the mean and standard deviations (SD) for the 10 sets of validation errors are given.  DeepKriging prediction is based on the MSE loss.  The baseline DNN includes only coordinates $\mathbf{s}$ in the features. The Kriging prediction uses an estimated exponential covariance function.}
	\vspace{2mm}
	\centering
	\begin{tabular}{c|c| c| c| c |c | c}
		\hline
		\multirow{2}{13mm}{Models}  &\multicolumn{2}{c|}{DeepKriging} & \multicolumn{2}{c|}{Baseline DNN} &\multicolumn{2}{c}{Kriging}   \\& Mean & SD & Mean & SD &  Mean & SD \\
		\hline
		\multicolumn{7}{l}{Training set}\\
		\hline
		RMSE ($\times10^{-3}$) & $\textbf{.585}$ &  $.869$ &  $4.255$ &  $2.384$ &5.969e-4&7.287e-5  \\
		MAPE &  $\textbf{1.269}$ &  $1.313$ &  $5.296$ &   $3.170$&5.749e-7&5.200e-8\\
		\hline
		\multicolumn{7}{l}{Testing set}\\
		\hline
		RMSE ($\times10^{-3}$) & $\textbf{3.466}$ &  $3.417$ &  $6.934$ &  $7.672$ &8.552&9.562  \\
		MAPE  &  $\textbf{5.330}$ &  $3.885$ &  $6.152$ &   $3.221$&0.007&0.005    \\
		\hline
	\end{tabular}
	\label{tab:2dsim}
\end{table}  

\section{Additional Simulations}
\subsection{Choice of basis functions}
As we have stated in the main manuscript, based on the KL theorem, the form of basis functions is not as important as the number of basis functions to approximate the spatial random effect $\nu(\mathbf{s})$. This can be supported by the additional simulations we conduct below.

\begin{figure}[!h] 
	\centering
	\includegraphics[width=0.75\linewidth]{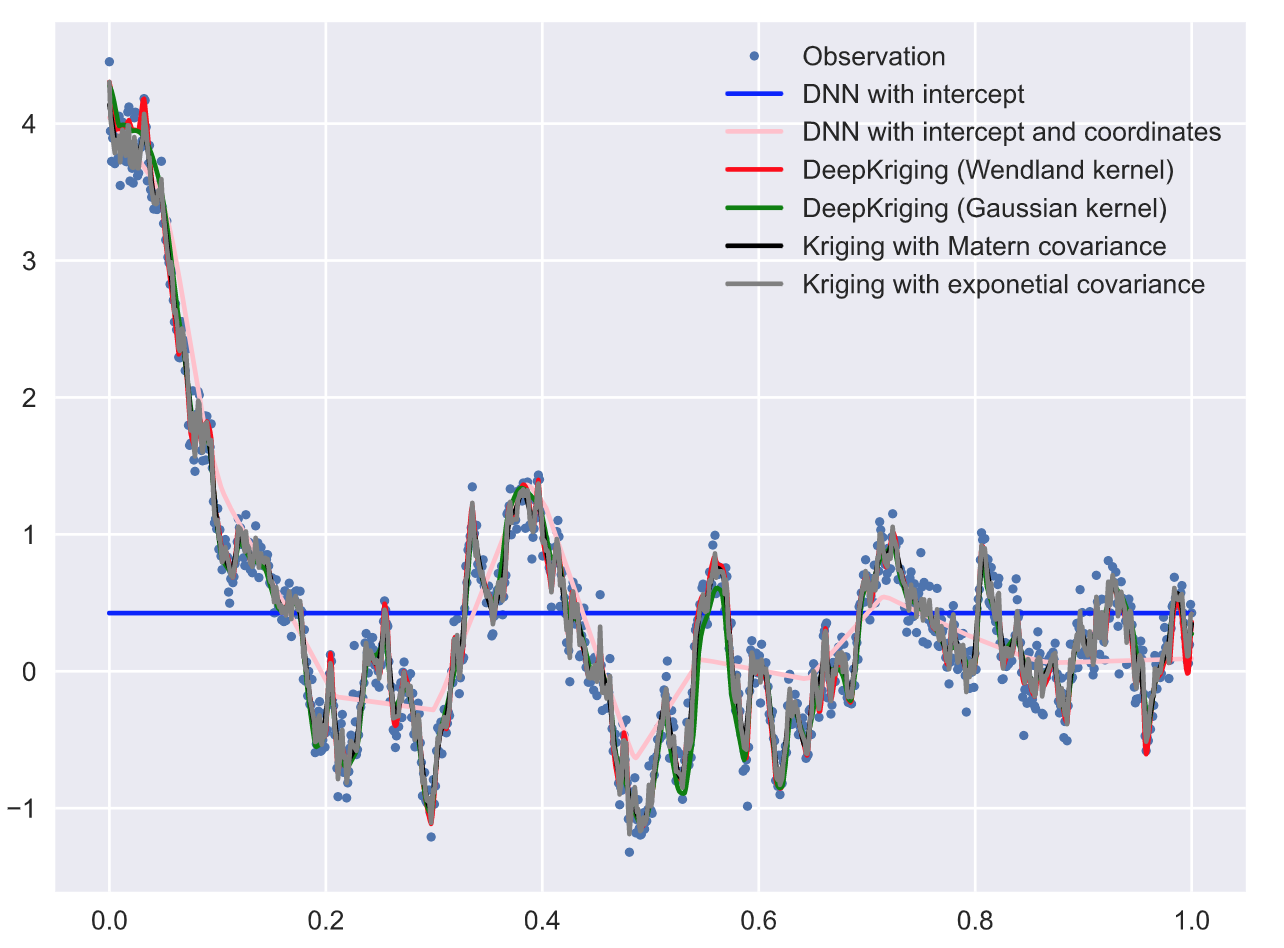}
	\vspace{-3mm}
	\caption{\baselineskip=16pt Prediction results for one simulated dataset generated from a GP. The blue dots are the simulated data (observations). The solid lines are the prediction curves from DNN with intercept only (blue line), DNN with intercept and coordinates (orange line), DeepKriging with Wenland kernel basis functions (red line), DeepKriging with Gaussian kernel basis functions (green line), Kriging with the true exponential covariance function (grey line), and Kriging with an estimated Mat{\'e}rn covariance function (black line), respectively. }
	\label{fig:1dsims2}
\end{figure}

In \cite{Nychka:15}, they stated that ``the choice of the Wendland family of RBFs is not crucial and other compactly supported, positive definite functions will work.'' They used compactly supported kernels so that the key matrices are sparse to improve the computational efficiency. Our DeepKriging method does not need matrix operations, hence in principle we could use any positive definite kernels. For the 1-D simulation study of Section~4.1, here we replace our original choice of the Wendland kernel with a Gaussian kernel. Specifically, at a certain level of resolution, let $\{\mathbf{u}_j\}$, $j=1,\ldots,m$, be a rectangular grid of points (i.e., the node points), and let $\theta$ be a scale parameter. The basis functions are then given by $\phi^{*}_j(\mathbf{s})=\phi(\|\mathbf{s}-\mathbf{u}_j\|/\theta)$, where $\phi(d)=\exp\{-d^2\}$. 
We use the same experimental designs for the 1-D simulation as before. Figure~\ref{fig:1dsims2} shows the prediction for one of the sample datasets using each of the six prediction methods. DeepKriging prediction with both basis function choices almost overlap with the optimal Kriging prediction. We then compare the RMSEs and MAPEs on both the training and testing datasets, as shown in Table~\ref{tab:train1D} below. The results for predictions other than DeepKriging II (with the new basis functions) are only slightly different from our original results in Table~\ref{tab:1D}, and the results are very close to each other between the two DeepKriging predictions. In addition, changing the basis functions do not change the conclusions overall; that is, for the testing set, the Kriging prediction with the true covariance function has the smallest RMSE as expected, and the performance of DeepKriging is comparable to the two Kriging predictions and outperforms the two naive DNN models. For the training set, both DeepKriging predictions are comparable to the optimal Kriging prediction, and they outperform the Kriging prediction with an estimated covariance function and the two naive DNN models in terms of both RMSE and MAPE.

\begin{figure}[h!] 
	\centering
	\includegraphics[width = .46\linewidth]{nonstat_fun_2d.pdf}
	\includegraphics[width = .53\linewidth]{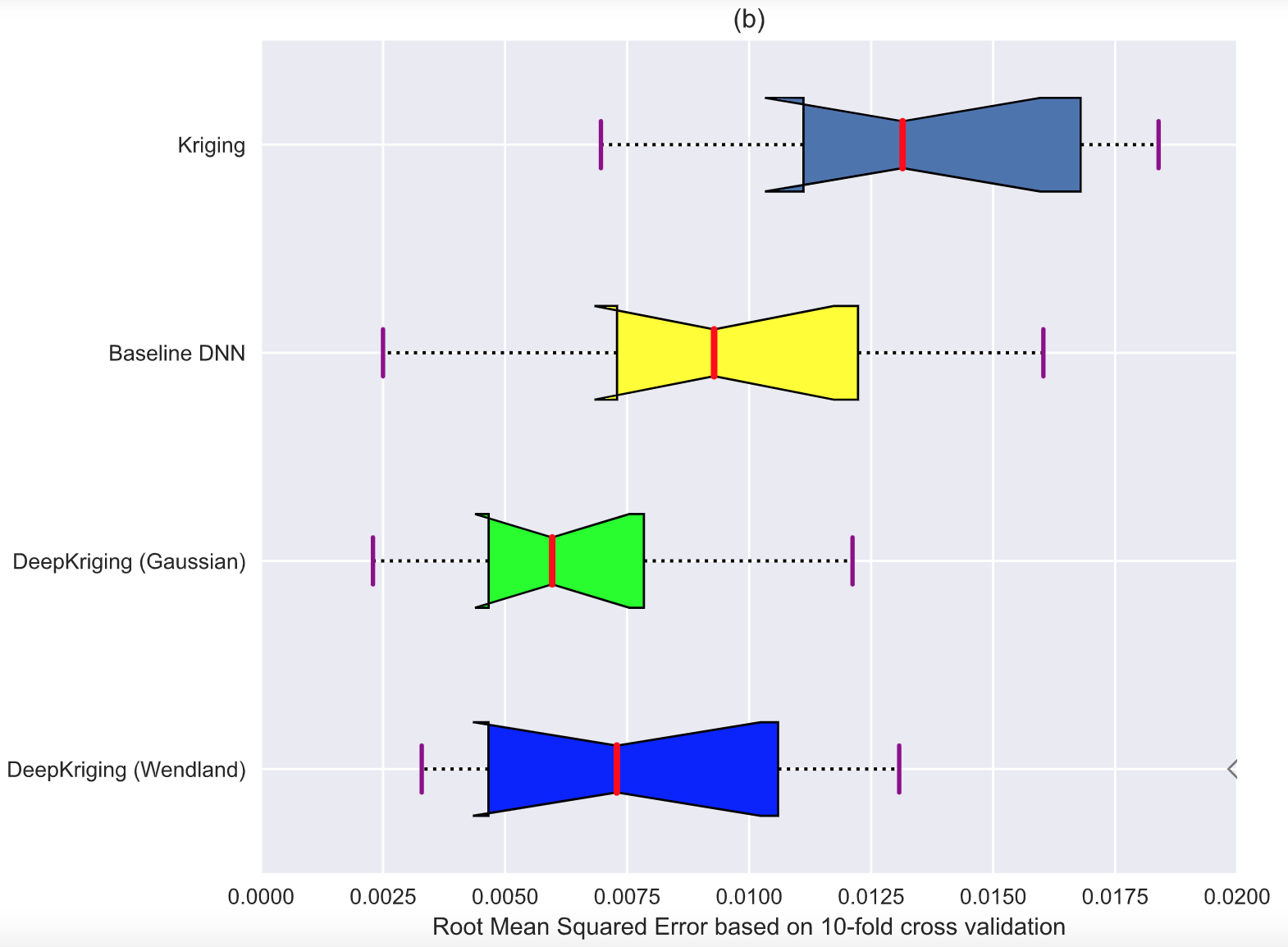}
	\vspace{-10mm}
	\caption{\baselineskip=19pt (a) Visualization of the simulated data generated from $Y(\mathbf{s}) = \sin\{30(\bar{s}-0.9)^4\}\cos\{2(\bar{s}-0.9)\}+(\bar{s}-0.9)/2$, where $\mathbf{s}=(s_x,s_y)^T\in [0,1]^2$ and $\bar{s}= (s_x+s_y)/2$. (b) Boxplots of the 10 RMSEs based on the 10-fold cross-validations from DeepKriging with Wendland kernel (blue), DeepKriging with Gaussian kernel (green), baseline DNN (yellow), and Kriging (gray).}
	\label{fig:2dsims2}
\end{figure}

Similarly for the 2-D simulation study in Section~4.2, replacing the radial basis functions using Wendland kernel with those using Gaussian kernel does not change the conclusion, if we compare Figure~\ref{fig:2dsims2}(b) with Figure~\ref{fig:2dsims}(b), and compare Table~\ref{tab:2dsim2} with Table~\ref{tab:2dsim}. Specifically, for the testing set, in terms of RMSE, both DeepKriging predictions significantly outperforms Kriging in terms of RMSEs and MAPEs. In addition, the baseline DNN is better than Kriging because the data are non-Gaussian and Kriging is no longer optimal. Moreover, the baseline DNN performs worse than DeepKriging as expected. The MAPE from DeepKriging is lower than the baseline DNN but higher than Kriging; this can happen since we are using MSE as the loss function in DeepKriging so it not necessarily possesses the lowest MAPE. For the training set, Kriging performs best in terms of both metrics since Kriging tends to underestimate such a variance, leading to a worse prediction on the testing dataset.

\begin{table}[!h]
	\caption{\baselineskip=16pt \label{tab:train1D}Mean and standard deviations (SD) of the root mean squared errors (RMSEs) and mean absolute percentage errors (MAPEs) on both the training and testing sets across the 100 datasets from the six predictions of a Gaussian process. Kriging I and II are the Kriging prediction with the true exponential covariance function and that with an estimated Mat{\'e}rn covariance function, respectively. DeepKriging I and II are the DeepKriging prediction with Wenland kernel and Gaussian kernel, respectively, based on the MSE loss. DNN I and DNN II are the DNN prediction with intercept only and that with intercept and coordinates, respectively.}
	\vspace{2mm}
	\centering
		\renewcommand{\arraystretch}{1.2}
	\resizebox{\textwidth}{!}{%
		\begin{tabular}{c|c| c| c| c| c| c |c | c| c | c| c | c}
			\hline
			\multirow{2}{13mm}{Models} &\multicolumn{2}{c|}{Kriging I} & \multicolumn{2}{c|}{Kriging II} &\multicolumn{2}{c|}{DeepKriging I} &\multicolumn{2}{c|}{DeepKriging II} &\multicolumn{2}{c|}{DNN I} &\multicolumn{2}{c}{DNN II} \\
			&Mean & SD & Mean & SD & Mean & SD & Mean & SD & Mean & SD & Mean & SD\\
			\hline
			\multicolumn{13}{l}{Training set}\\
			\hline
			RMSE & $\textbf{.063}$ &  $.002$ &  $.125$ &  $.006$ &.087&.011&.120&.008&.887&.183&.243&.025\\
			\hline
			MAPE &  $\textbf{.217}$ &  $.128$ &  $.456$ &   $.318$&.274&.189&.450&.453&4.030&4.725&.915&.921\\
			\hline
			\multicolumn{13}{l}{Testing set}\\
			\hline
			RMSE & $\textbf{.160}$ &  $.008$ &  $.165$ &  $.008$ &.197&.018&.187&.015&.884&.186&.254&.029\\
			\hline
			MAPE &  $\textbf{.573}$ &  $.541$ &  $.603$ &   $.583$&.676&.639&.676&.635&3.495&2.747&.880&.781\\
			\hline
		\end{tabular}
	}
\end{table}

\begin{table}[h!]
	\caption{\baselineskip=16pt Model performance for both the training and testing sets based on 10-fold cross-validation. Data are generated from $Y(\mathbf{s}) = \sin\{30(\bar{s}-0.9)^4\}\cos\{2(\bar{s}-0.9)\}+(\bar{s}-0.9)/2$, where $\mathbf{s}=(s_x,s_y)^T\in [0,1]^2$ and $\bar{s}= (s_x+s_y)/2$. RMSEs and MAPEs are computed, and the mean and standard deviations (SD) for the 10 sets of validation errors are given.  DeepKriging I and II are the DeepKriging predictions with Wenland kernel and Gaussian kernel, respectively, based on the MSE loss.  The baseline DNN includes only coordinates $\mathbf{s}$ in the features. The Kriging prediction uses an estimated exponential covariance function.}
	\vspace{2mm}
	\centering
	\resizebox{\textwidth}{!}{%
		\begin{tabular}{c|c| c| c| c| c| c |c | c}
			\hline
			\multirow{2}{13mm}{Models} &\multicolumn{2}{c|}{DeepKriging I} &\multicolumn{2}{c|}{DeepKriging II} & \multicolumn{2}{c|}{Baseline DNN} &\multicolumn{2}{c}{Kriging}   \\& Mean & SD & Mean & SD &  Mean & SD & Mean & SD\\
			\hline
			\multicolumn{9}{l}{Training set}\\
			\hline
			RMSE ($\times10^{-3}$) & $\textbf{.526}$ &  $.062$ &  $\textbf{1.382}$ &  $.082$ &6.946&3.946&1.757e-4&2.399e-4  \\
			MAPE  &  $\textbf{.732}$ &  $.838$ &  $\textbf{1.386}$ &   $.622$&9.634&6.389&5.152e-7&6.624e-8     \\
			\hline
			\multicolumn{9}{l}{Testing set}\\
			\hline
			RMSE ($\times10^{-3}$) & $\textbf{8.431}$ &  $4.927$ &  $\textbf{6.578}$ &  $3.065$ &9.987&5.375&15.135&7.804  \\
			MAPE  &  $\textbf{10.520}$ &  $5.397$ &  $\textbf{5.798}$ &   $2.361$&10.875&7.494&.098&.088    \\
			\hline
		\end{tabular}
	}
	\label{tab:2dsim2}
\end{table}  

\subsection{DeepKriging is non-linear}\label{sec:nonlinear}
\begin{figure}[h] 
	\centering
	\includegraphics[width = .49\linewidth]{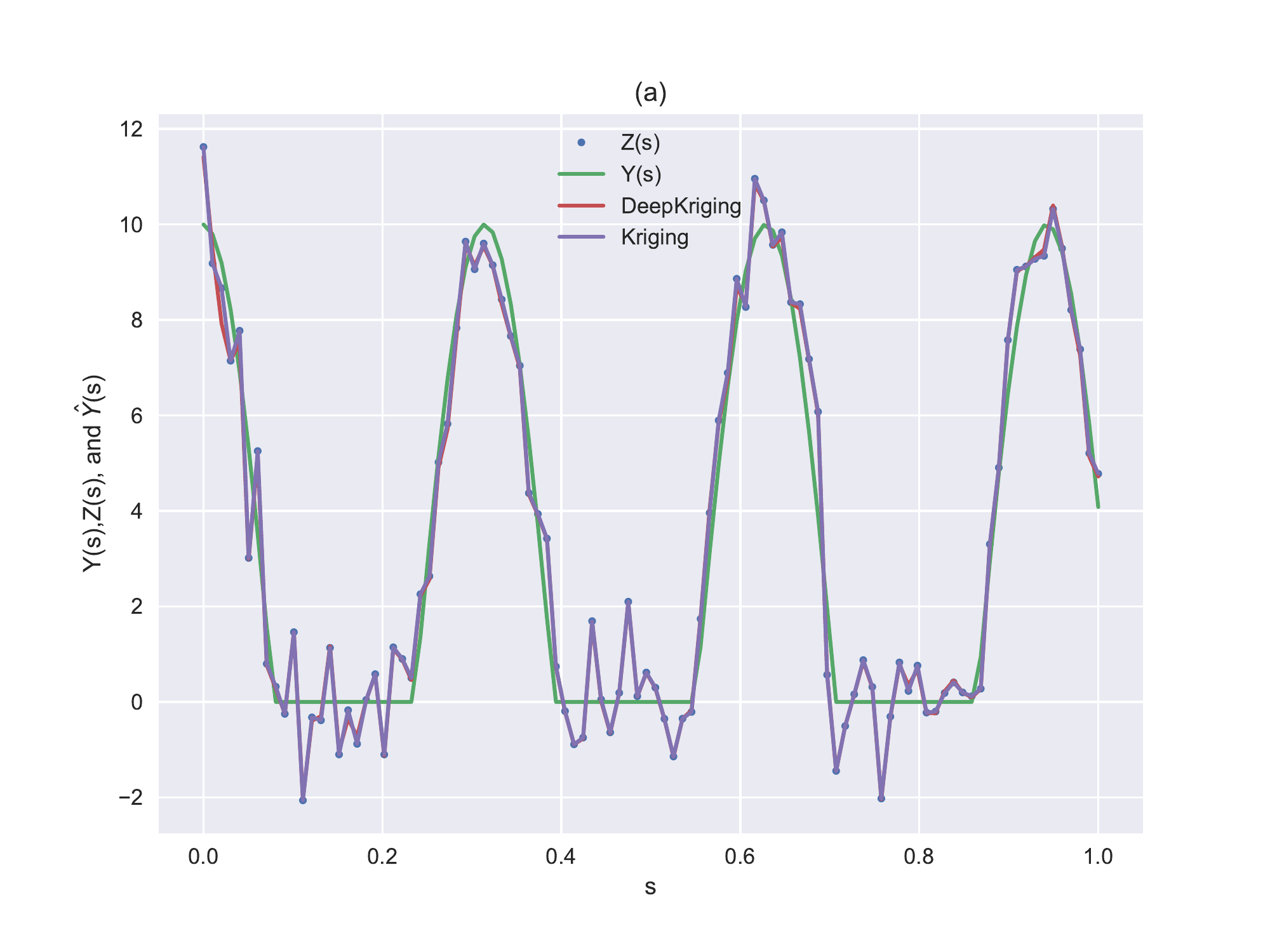}
	\includegraphics[width = .49\linewidth]{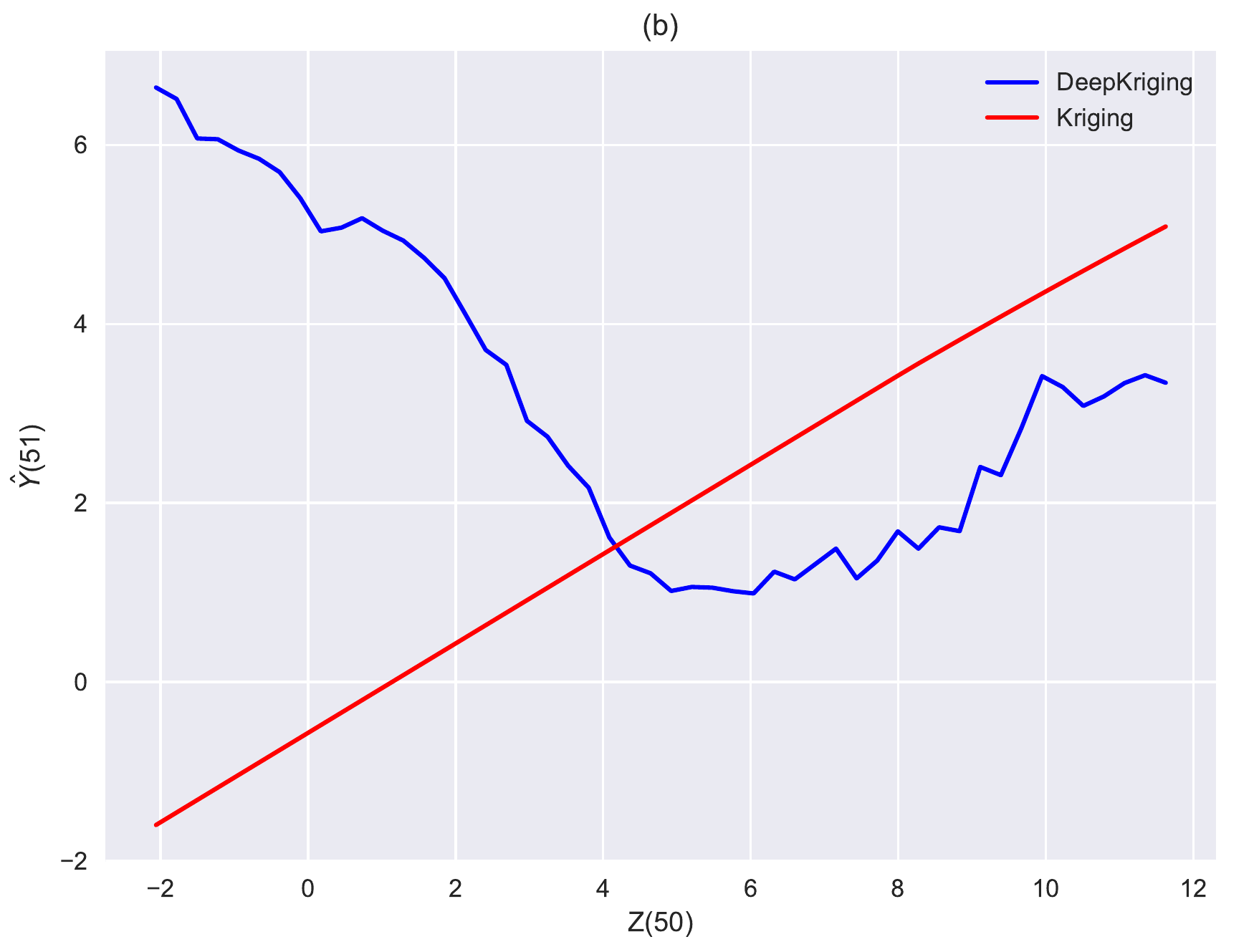}
	\vspace*{-5mm}
	\caption{\baselineskip=20pt (a) Visualization of the simulated data and predictions.  The observations (blue dots) z(s) are generated by the the signal (green line) $Y(s)\mathbbm{1}_{\{Y(s)>0\}}+\varepsilon(s)$, where $Y(s)=10\cos(20s)$, with a standard Gaussian noise. The predictions of Kriging (purple line) and DeepKriging (red line) are almost identical. (b) The relationship and sensitivity of $z(50)$ on $\hat{Y}(51)$ from Kriging (red line) and DeepKriging (blue line) prediction.s The x-axis shows 50 different values of $z(50)$ in the model training and the y-axis shows 50 predicted values of $Y(51)$ provided that $z(51)$ is not used.}
	\label{fig:1dlinear}
\end{figure}

It is important to investigate how the DeepKriging prediction is linked with the observations. We generate 100 observations generated by $z(s)=Y(s)\mathbbm{1}_{\{Y(s)>0\}}+\varepsilon(s)$, where $Y(s)=10\cos(20s)$, the coordinates $s$ are regularly located in $[0,1]$, and $\varepsilon(s)\overset{iid}{\sim} N(0,1)$. The simulated data $z(s)$ and signal $Y(s)$ are shown in Figure \ref{fig:1dlinear}(a). The figure also shows the prediction results from both Kriging and DeepKriging, which are almost identical. 

To examine the non-linear relationship between a training observation and the prediction, we replace the observation $z(s)$ at $s=50$ by $m=50$ different values and drop the observation $z(s+1)$ in the model training, and obtain the prediction $\hat{Y}(s+1)$ by Kriging or DeepKriging. By doing so, we are able to test the relationship and sensitivity of $z(50)$ on $\hat{Y}(51)$ for both methods. The results in Figure \ref{fig:1dlinear}(b) show that the Kriging prediction is linear in observations, while the DeepKriging provides an obviously nonlinear prediction in observations. This simulation study shows that although the Kriging and DeepKriging predictions are almost identical, the underlying relationships between the prediction and observations are totally different.

\vspace*{-5mm}
\subsection{Computational time of DeepKriging}
\begin{figure}[h!]
	\centering
	\includegraphics[width = .7\linewidth]{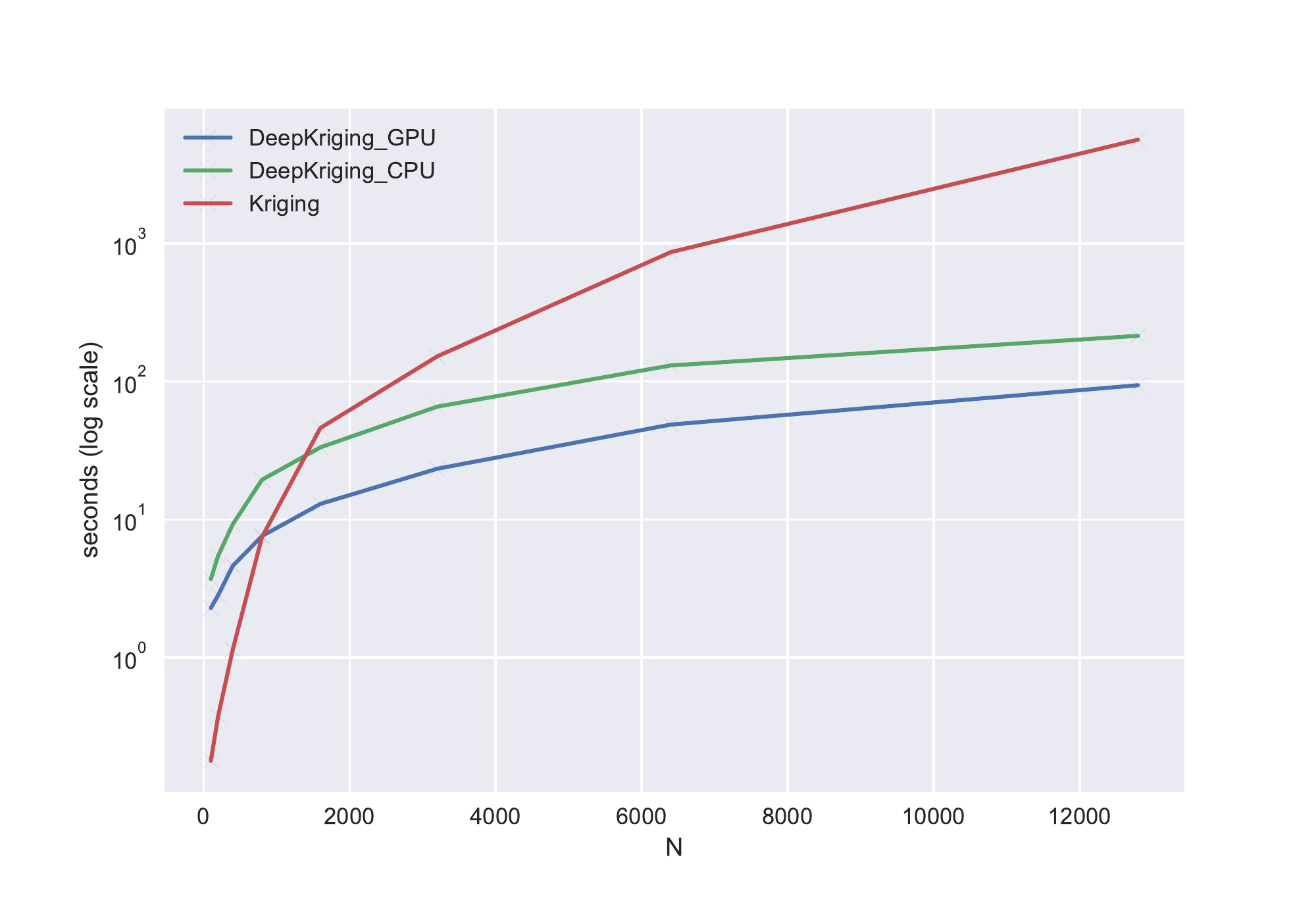}
	\vspace*{-5mm}
	\caption{\baselineskip=20pt Computational time in seconds ($\log_{10}$ scale) of DeepKriging and Kriging for different sample sizes $N$. The green line is the computational time of Kriging. The blue and green lines are the computational time of DeepKriging based on a Tesla P100 GPU and a 2.5 GHz Intel Core i7 CPU, respectively.}
	\label{fig:runtime}
\end{figure}

Based on the same simulation setting in Section~\ref{sec:nonlinear}, we investigate the computational time of DeepKriging compared to Kriging with different sample sizes $N$. 
Figure \ref{fig:runtime} shows the results. Although Kriging is faster for small sample sizes ($N<1,500$), DeepKriging is much more scalable when the sample size increases. For example, when $N=12,800$, which is the largest sample size we have considered, it takes more than 1.5 hours ($5,663$ seconds) to train a Kriging model, but DeepKriging only costs 3.5 minutes ($214$ seconds) without GPU acceleration and 1.5 ($94$ seconds) minutes with a Tesla P100 GPU. Note that the GPU we use is freely accessed in Kaggle. For a more powerful GPU, the computational cost will be further reduced.

\section{Distribution prediction and uncertainty quantification of DeepKriging}\label{sec:densityregression}

An ideal spatial prediction not only provides point prediction, but also distributional information such as quantiles or the density function to quantify uncertainties, risks and extreme values \citep{diggle2007spatial}. Traditional DNNs cannot provide the uncertainty information at the predicted spatial locations. Recently, several methods have been proposed to overcome this problem by predicting the entire probabilistic distribution.  

\citet{gal2016dropout} and \citet{posch2019variational} applied Bayesian inference methodologies to neural networks to predict uncertainties via the posterior distribution. However, these methods cannot be applied directly to spatial data. Some studies try to combine deep learning and GPs which can be potentially applied in spatial prediction \citep{damianou2013deep, lee2017deep, kumar2018deep, zammit2019deep, blomqvist2019deep}. These methods are often called deep Gaussian processes that characterize deep learning in the framework of GPs via Bayesian learning \citep{neal2012bayesian}. Based on the correspondence between an infinitely wide DeepKriging and GPs as illustrated in Section 3.3 in the main manuscript, we can potentially compute the covariance function for the induced GPs and then use the resulting GPs to perform Bayesian inference for wide DeepKriging networks. However, it is infeasible to train an infinitely wide DNN in practice, and \citet{lee2017deep} found that the test performance increases as finite-width trained networks are made wider and more similar to a GP, and thus that GP predictions typically outperform those of finite-width networks. The related computation burden is another issue for applications in large datasets.

The Bayesian methods mentioned above require configurations on the prior distributions for hyper-parameters. \citet{li2019deep} proposed a completely distribution-free method for density estimation called deep distribution regression (DDR), where the density was approximated using histograms with discrete bins, and the discretized density was estimated by multi-class classification using neural networks. Although the method was designed for a regression problem, we can extend the idea to our DeepKriging framework for predictive distribution estimation. The idea is straightforward and the method is easy to implement. We call the method deep distribution spatial prediction (DDSP), as illustrated below.

\subsection{Deep Distribution Spatial Regression}
We quantify the uncertainty of DeepKriging prediction by the conditional distribution of ${Y}(\mathbf{s}_0)|\mathbf{z}, \mathbf{x}_{\phi}(\mathbf{s}_0)$ at an unobserved location $\mathbf{s}_0$. We denote the probability density function (PDF) as $f_\mathrm{UQ}(y|x)$, whose support is $[l, u]$, with $l$ and $u$ being the lower and upper bound, respectively. The interval can be partitioned into $M+1$ bins, $T_m=[c_{m-1},c_m)$, by $M$ cut-points such that $c_0<c_1<\dots<c_M<c_{M+1}$, where $c_0=l$ and $c_{M+1}=u$. Let $|T_m|$ be the length of the $m$-th bin and $p_m(\mathbf{s}_0)=\mathbb{P}\{Y(\mathbf{s}_0)\in T_m| \mathbf{z}, \mathbf{x}_{\phi}(\mathbf{s}_0)\}$ be the conditional probability that $Y(\mathbf{s}_0)$ falls into the $m$-th bin. Then $f_\mathrm{UQ}(y|x)$ is approximated by $M+1$ constant functions, ${p_m(\mathbf{s}_0)}/{|T_m|},m=1,\dots,M+1$. The density prediction is specified as:
\begin{equation}
	\hat f_\mathrm{UQ}(y|x)=\sum_{m=1}^{M+1}\frac{\widehat{p}_m(\mathbf{s}_0)}{|T_m|}\mathbbm{1}\{y\in T_m\}.
	\label{eq:DDR}
\end{equation} 

The crucial step in Equation \eqref{eq:DDR} is to estimate the bin probabilities $\{\widehat{p}_1(\mathbf{s}_0),\dots, \widehat{p}_{M+1}(\mathbf{s}_0)\}$ using a classification model, so that neural networks can be applied. \citet{li2019deep} suggested different ways for loss functions and bin partitioning. The most natural way is to use multi-class classification with fixed bins. In the output layer, a softmax function is applied to ensure that $\{\widehat{p}_1(\mathbf{s}_0),\dots, \widehat{p}_{M+1}(\mathbf{s}_0)\}$ constitutes a valid probability vector. The loss function for this case is the negative multinomial log-likelihood function, which is equivalent to the multi-class cross entropy loss: $\sum_{n=1}^N\sum_{m=1}^{M+1}\mathbbm{1}\{z(\mathbf{s}_n)\in T_m\}\log\{p_m(\mathbf{s}_n)\}$. 


In practice, however, the estimation of a continuous density function by the multi-class fixed classification is sensitive to the choice of bins. Therefore, for DDSP in DeepKriging, we use the improved option with joint binary cross entropy loss function (JBCE) and ensembles. The JBCE function is specified as 
\begin{equation}
	\rm{JBCE} = \sum_{n=1}^N\sum_{m=1}^{M}\left[\mathbbm{1}\{z(\mathbf{s}_n)\leq c_m\}\log\{F(c_m;\mathbf{s}_n)\}+\mathbbm{1}\{z(\mathbf{s}_n)>c_m\}\log\{1-F(c_m;\mathbf{s}_n)\}\right],
	\label{eq:lik}
\end{equation}
where $F(c_m;\mathbf{s}_n)=\sum_{i=1}^mp_i(\mathbf{s}_n)$ and $c_m, m=1,\dots,M$ are the $M$ cut-points. 

The DDSP may provide a non-smooth density. Thus, an ensemble method can be applied for further adjustment by fitting $I$ independent classifications and computing the average of classifiers. Algorithm \ref{alg:ensemble} provides the procedure of obtaining density prediction of DeepKriging with the ensemble of random partitioning:
\begin{spacing}{1.5}
\begin{algorithm}[h]
	\SetAlgoLined
	\For{i =1:I}{
		Draw $M$ cut-points from Uniform(0,1)\;
		Sort the cut-points as $c_{i1},\dots,c_{iM}$\;
		Assign $Y(\mathbf{s}_0)$\ to one of the $M$ bins, where the $m-th$ bin is $T_{im}=[c_{i,m-1},c_{im})$\;
		Train the classifier to estimate the probabilities $\widehat{p}_{i1}(\mathbf{s}_0),\dots, \widehat{p}_{i,M+1}(\mathbf{s}_0)$\;}
	\KwResult{$\hat f_\mathrm{UQ}(y|x)=\sum_{i=1}^{I}\sum_{m=1}^{M+1}\frac{\widehat{p}_{im}(\mathbf{s}_0)}{|T_{im}|}\mathbbm{1}\{y\in T_{im}\}$}
	\caption{The algorithm of density prediction of DeepKriging with ensemble random partitioning. We ensemble I independent classifications, for each of which $M$ bins are chosen randomly.}
	\label{alg:ensemble}
\end{algorithm}
\end{spacing}

Note that although the density regression provides an uncertainty quantification without any distribution assumption, it may bring new uncertainties from the neural networks. Therefore, the GP process representation may be a better way for uncertainties if it is quantified by the standard deviation, even though the computational burden will increase.

In addition, the density prediction requires further computations if we apply ensembles, where each step in the ensemble will implement classification separately with a new choice of random cut-points. However, the main objective of the ensemble in the density prediction is to obtain a smooth density. If the research goal is to get uncertainties using quantiles without the density curve, then the ensemble is not always needed. Another hyperparameter we need to specify in the density prediction is the bin size $M+1$. Typically, large $M$ will provide a more complex pattern of the density, and small $M$ will give a more stable estimation. In practice, we can follow the Freedman--Diaconis rule \citep{freedman1981histogram} as a robust estimation in the histogram approximation such that $M = \left \lfloor \{\max(\mathbf{z})-\min(\mathbf{z})\}\frac{\sqrt[3]{N}}{2\times\mathrm{IQR}(\mathbf{z})}\right \rfloor$, where IQR($\mathbf{z}$) is the interquartile range of the observations.

\subsection{Uncertainty Quantification: A Simulation Study}
The simulated data is generated by a 1-D Gaussian mixture model: 
\[z(s) = \{\sin(5{s}) + 0.7 + \tau_1(s)\}\pi(s) + \{2\sin(8{s}) + \tau_2(s))\}(1 - \pi(s)),\]
 where $\tau_1(s)\overset{iid}{\sim} N(0,0.2^2)$, $\tau_2(s)\overset{iid}{\sim} N(0,0.3^2)$, $\pi(s)\overset{iid}{\sim} \rm{Bernoulli}(0.5)$, and $\tau_2(s)$, $\tau_2(s)$, and $\pi(s)$ are mutually independent.  Observations are drawn from $2,500$ regularly spaced locations in $[0,1]$, out of which $2,000$ are training samples and $500$ are testing samples.

\begin{figure}[t] 
	\centering
	\includegraphics[width = .75\linewidth]{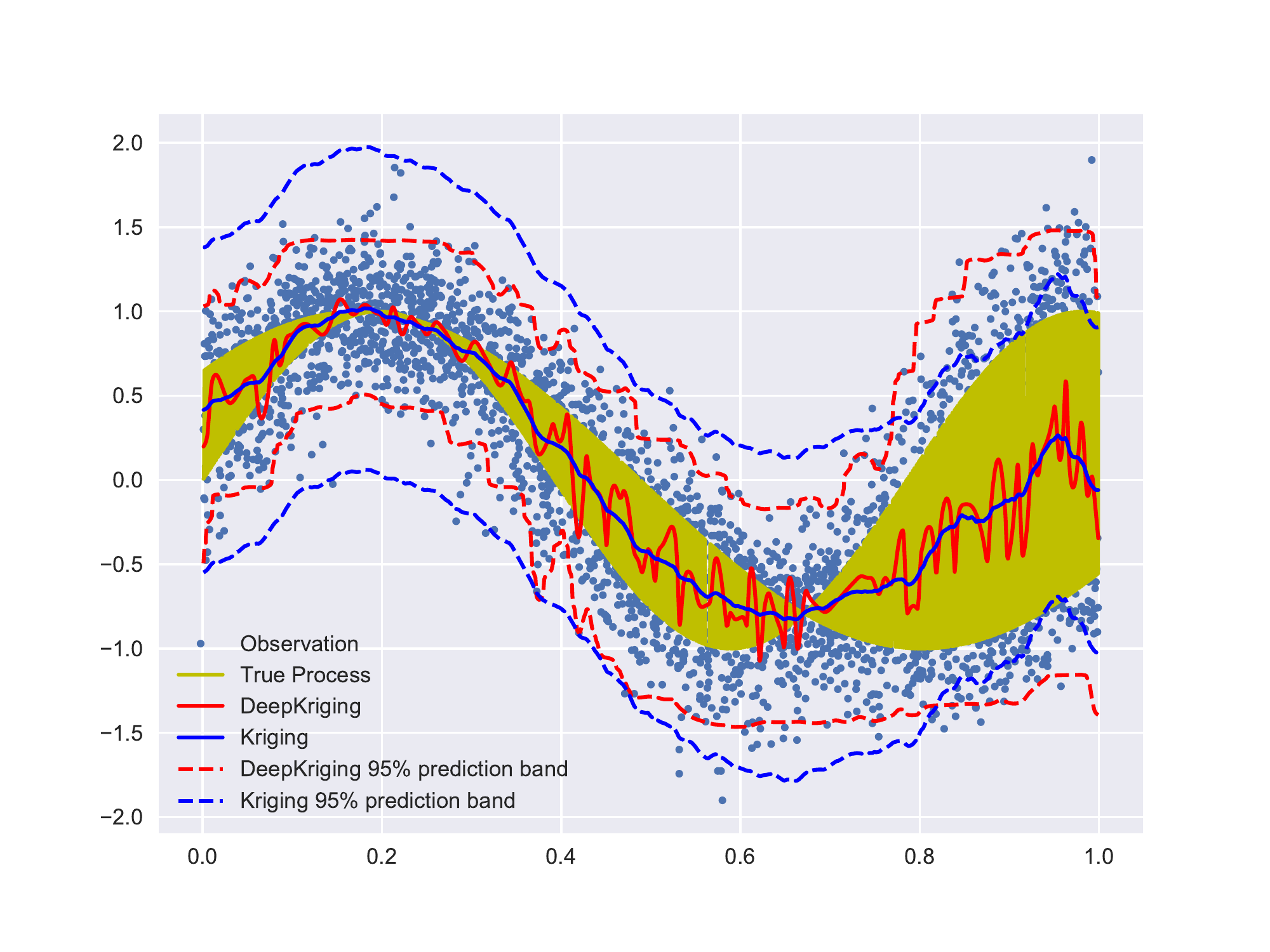}
	\vspace*{-3mm}
	\caption{\baselineskip=20pt Prediction results for a Gaussian mixture model. The blue dots are the observations simulated from the Gaussian mixture model. The solid lines are the true process (yellow) and the predictions from DeepKriging (red) and Kriging (blue), respectively. The dashed lines represent the 95\% prediction bands from DeepKriging (red) and Kriging (blue), respectively.}
	\label{fig:1dmix}
\end{figure}

Figure \ref{fig:1dmix} shows the prediction results on both training and testing samples. The advantage of DeepKriging is obvious for the uncertainty quantification. The process is heteroscedastic with an obviously large variance when $s$ is larger than $0.8$. The heteroscedasticity is well captured by the DeepKriging but missed by the Kriging. Moreover, the prediction band of the Kriging is too narrow for large $s$ and two large for small $s$.

Numerical evaluations on testing samples are further used for comparison based on the root mean squared error (RMSE) and average quantile loss (AQTL). AQTL is defined by
$\rm{AQTL}=\sum_{t=1}^{99}\sum_{n=1}^N(\{z(s_n)-\widehat{Q}(t/100)\}[t/100-\mathbbm{1}\{z({s}_n)\leq \widehat{Q}(t/100)\}])$, where $\widehat{Q}(t/100)$ is the estimated $t$-th percentile. The results show that the RMSE of the Kriging (0.490) is comparable to DeepKriging (0.485). But, in terms of the quantile loss, the AQTL of the Kriging (60.47) is much larger than that of DeepKriging (0.12), which further implies that Kriging is only a good estimation in terms of the mean squared error which reflects the mean and variance, but cannot reflect other important properties such as quantiles. 

\section{Source codes and data}
\begin{itemize}
	\item The source codes can be found in the following Github repository: \\
	\href{https://github.com/aleksada/DeepKriging}{https://github.com/aleksada/DeepKriging}. 
	
	\item The methods are implemented in Keras, a library of both Python and R. Most of the codes for this work are written in Python. We also provide some simple examples of DeepKriging using Keras in R.
	
	\item The PM2.5 station data from AQS used in the application can be found in 
	\href{https://aqs.epa.gov/aqsweb/airdata/download_files.html}{https://aqs.epa.gov/aqsweb/airdata/download\_files.html}.
	
	\item The meteorological data from NARR used in the application can be found in 
	\href{https://psl.noaa.gov/data/gridded/data.narr.html}{https://psl.noaa.gov/data/gridded/data.narr.html}.
\end{itemize}

\renewcommand\bibname{\Large \bf References}

\begin{thebibliography}{}
	
	\bibitem[Adgate et~al., 2002]{adgate2002spatial}
	Adgate, J.~L., Ramachandran, G., Pratt, G., Waller, L., and Sexton, K. (2002).
	\newblock Spatial and temporal variability in outdoor, indoor, and personal
	{PM2.5} exposure.
	\newblock {\em Atmospheric Environment}, 36(20):3255--3265.
	
	\bibitem[Adler, 2010]{adler2010geometry}
	Adler, R.~J. (2010).
	\newblock {\em The Geometry of Random Fields}.
	\newblock SIAM.
	
	\bibitem[Anselin, 2001]{anselin2001spatial}
	Anselin, L. (2001).
	\newblock Spatial econometrics.
	\newblock {\em A Companion to Theoretical Econometrics}, 310330.
	
	\bibitem[Austin, 2002]{austin2002spatial}
	Austin, M. (2002).
	\newblock Spatial prediction of species distribution: an interface between
	ecological theory and statistical modeling.
	\newblock {\em Ecological Modeling}, 157(2-3):101--118.
	
	\bibitem[Banerjee et~al., 2010]{banerjee2010hierarchical}
	Banerjee, S., Finley, A.~O., Waldmann, P., and Ericsson, T. (2010).
	\newblock Hierarchical spatial process models for multiple traits in large
	genetic trials.
	\newblock {\em Journal of the American Statistical Association},
	105(490):506--521.
	
	\bibitem[Banerjee et~al., 2008]{banerjee2008gaussian}
	Banerjee, S., Gelfand, A.~E., Finley, A.~O., and Sang, H. (2008).
	\newblock Gaussian predictive process models for large spatial data sets.
	\newblock {\em Journal of the Royal Statistical Society: Series B (Statistical
		Methodology)}, 70(4):825--848.
	
	\bibitem[Cheng et~al., 2016]{cheng2016wide}
	Cheng, H.-T., Koc, L., Harmsen, J., Shaked, T., Chandra, T., Aradhye, H.,
	Anderson, G., Corrado, G., Chai, W., Ispir, M., et~al. (2016).
	\newblock Wide \& deep learning for recommender systems.
	\newblock In {\em Proceedings of the 1st workshop on deep learning for
		recommender systems}, pages 7--10.
	
	\bibitem[Cho and Saul, 2009]{cho2009kernel}
	Cho, Y. and Saul, L.~K. (2009).
	\newblock Kernel methods for deep learning.
	\newblock In {\em Advances in neural information processing systems}, pages
	342--350.
	
	\bibitem[Cracknell and Reading, 2014]{cracknell2014geological}
	Cracknell, M.~J. and Reading, A.~M. (2014).
	\newblock Geological mapping using remote sensing data: A comparison of five
	machine learning algorithms, their response to variations in the spatial
	distribution of training data and the use of explicit spatial information.
	\newblock {\em Computers \& Geosciences}, 63:22--33.
	
	\bibitem[Cressie, 1990]{cressie1990origins}
	Cressie, N. (1990).
	\newblock The origins of kriging.
	\newblock {\em Mathematical Geology}, 22(3):239--252.
	
	\bibitem[Cressie, 2015]{cressie2015statistics}
	Cressie, N. (2015).
	\newblock {\em Statistics for Spatial Data}.
	\newblock John Wiley \& Sons.
	
	\bibitem[Cressie and Johannesson, 2008]{cressie2008fixed}
	Cressie, N. and Johannesson, G. (2008).
	\newblock Fixed rank kriging for very large spatial data sets.
	\newblock {\em Journal of the Royal Statistical Society: Series B},
	70(1):209--226.
	
	\bibitem[Cs{\'a}ji, 2001]{csaji2001approximation}
	Cs{\'a}ji, B.~C. (2001).
	\newblock Approximation with artificial neural networks.
	\newblock {\em Faculty of Sciences, Etvs Lornd University, Hungary}, 24:48.
	
	\bibitem[DeGroot, 2005]{degroot2005optimal}
	DeGroot, M.~H. (2005).
	\newblock {\em Optimal Statistical Decisions}, volume~82.
	\newblock John Wiley \& Sons.
	
	\bibitem[Di et~al., 2016]{di2016assessing}
	Di, Q., Kloog, I., Koutrakis, P., Lyapustin, A., Wang, Y., and Schwartz, J.
	(2016).
	\newblock Assessing {PM2.5} exposures with high spatiotemporal resolution
	across the continental {United States}.
	\newblock {\em Environmental Science \& Technology}, 50(9):4712--4721.
	
	\bibitem[Dubrule, 1983]{dubrule1983two}
	Dubrule, O. (1983).
	\newblock Two methods with different objectives: splines and kriging.
	\newblock {\em Journal of the International Association for Mathematical
		Geology}, 15(2):245--257.
	
	\bibitem[Dubrule, 1984]{dubrule1984comparing}
	Dubrule, O. (1984).
	\newblock Comparing splines and kriging.
	\newblock {\em Computers \& Geosciences}, 10(2-3):327--338.
	
	\bibitem[EPA, 2012]{standard}
	EPA, U. (2012).
	\newblock National ambient air quality standards (NAAQS).
	\newblock \url{https://www.epa.gov/criteria-air-pollutants/naaqs-table}.
	\newblock Date accessed: [Dec 15, 2019].
	
	\bibitem[Fan et~al., 2019]{fan2019selective}
	Fan, J., Ma, C., and Zhong, Y. (2019).
	\newblock A selective overview of deep learning.
	\newblock {\em arXiv preprint arXiv:1904.05526}.
	
	\bibitem[Fong and Tyler, 2021]{fong2021machine}
	Fong, C. and Tyler, M. (2021).
	\newblock Machine learning predictions as regression covariates.
	\newblock {\em Political Analysis}, 29(4):467--484.
	
	\bibitem[Franchi et~al., 2018]{franchi2018supervised}
	Franchi, G., Yao, A., and Kolb, A. (2018).
	\newblock Supervised deep kriging for single-image super-resolution.
	\newblock In {\em German Conference on Pattern Recognition}, pages 638--649.
	Springer.
	
	\bibitem[Friedman et~al., 2001]{friedman2001elements}
	Friedman, J., Hastie, T., and Tibshirani, R. (2001).
	\newblock {\em The Elements of Statistical Learning}.
	\newblock Springer.
	
	\bibitem[Fuentes, 2002]{fuentes2002spectral}
	Fuentes, M. (2002).
	\newblock Spectral methods for nonstationary spatial processes.
	\newblock {\em Biometrika}, 89(1):197--210.
	
	\bibitem[Goodfellow et~al., 2016]{goodfellow2016deep}
	Goodfellow, I., Bengio, Y., and Courville, A. (2016).
	\newblock {\em Deep Learning}.
	\newblock MIT Press.
	
	\bibitem[Heaton et~al., 2019]{heaton2019case}
	Heaton, M.~J., Datta, A., Finley, A.~O., Furrer, R., Guinness, J., Guhaniyogi,
	R., Gerber, F., Gramacy, R.~B., Hammerling, D., Katzfuss, M., et~al. (2019).
	\newblock A case study competition among methods for analyzing large spatial
	data.
	\newblock {\em Journal of Agricultural, Biological and Environmental
		Statistics}, 24(3):398--425.
	
	\bibitem[Hennessey~Jr, 1977]{hennessey1977some}
	Hennessey~Jr, J.~P. (1977).
	\newblock Some aspects of wind power statistics.
	\newblock {\em Journal of Applied Meteorology}, 16(2):119--128.
	
	\bibitem[Higdon, 2002]{higdon2002space}
	Higdon, D. (2002).
	\newblock Space and space-time modeling using process convolutions.
	\newblock In {\em Quantitative methods for current environmental issues}, pages
	37--56. Springer.
	
	\bibitem[Imai and Khanna, 2016]{imai2016improving}
	Imai, K. and Khanna, K. (2016).
	\newblock Improving ecological inference by predicting individual ethnicity
	from voter registration records.
	\newblock {\em Political Analysis}, 24(2):263--272.
	
	\bibitem[Jacob and Winner, 2009]{jacob2009effect}
	Jacob, D.~J. and Winner, D.~A. (2009).
	\newblock Effect of climate change on air quality.
	\newblock {\em Atmospheric Environment}, 43(1):51--63.
	
	\bibitem[Katzfuss, 2017]{katzfuss2017multi}
	Katzfuss, M. (2017).
	\newblock A multi-resolution approximation for massive spatial datasets.
	\newblock {\em Journal of the American Statistical Association},
	112(517):201--214.
	
	\bibitem[Kleiber and Nychka, 2015]{kleiber2015equivalent}
	Kleiber, W. and Nychka, D.~W. (2015).
	\newblock Equivalent kriging.
	\newblock {\em Spatial Statistics}, 12:31--49.
	
	\bibitem[Krizhevsky et~al., 2012]{krizhevsky2012imagenet}
	Krizhevsky, A., Sutskever, I., and Hinton, G.~E. (2012).
	\newblock {ImageNet} classification with deep convolutional neural networks.
	\newblock In {\em Advances in Neural Information Processing Systems}, pages
	1097--1105.
	
	\bibitem[LeCun et~al., 2015]{lecun2015deep}
	LeCun, Y., Bengio, Y., and Hinton, G. (2015).
	\newblock Deep learning.
	\newblock {\em Nature}, 521(7553):436--444.
	
	\bibitem[Lee et~al., 2018]{lee2017deep}
	Lee, J., Sohl-Dickstein, J., Pennington, J., Novak, R., Schoenholz, S., and Bahri, Y. (2018).
	\newblock Deep neural networks as {Gaussian} processes.
	\newblock {\em International Conference on Learning Representations}.
	
	\bibitem[Lehmann and Casella, 2006]{lehmann2006theory}
	Lehmann, E.~L. and Casella, G. (2006).
	\newblock {\em Theory of point estimation}.
	\newblock Springer Science \& Business Media.
	
	\bibitem[Li and Sun, 2019]{Yuxiao2018}
	Li, Y. and Sun, Y. (2019).
	\newblock Efficient estimation of non-stationary spatial covariance functions
	with application to high-resolution climate model emulation.
	\newblock {\em Statistica Sinica}, 29(3):1209--1231.
	
	\bibitem[Matheron, 1963]{matheron1963principles}
	Matheron, G. (1963).
	\newblock Principles of geostatistics.
	\newblock {\em Economic Geology}, 58(8):1246--1266.
	
	\bibitem[Matheron, 1981]{matheron1981splines}
	Matheron, G. (1981).
	\newblock Splines and kriging: their formal equivalence.
	\newblock {\em Down-to-Earth-Statistics: Solutions Looking for Geological
		Problems}, pages 77--95.
	
	\bibitem[Najafabadi et~al., 2015]{najafabadi2015deep}
	Najafabadi, M.~M., Villanustre, F., Khoshgoftaar, T.~M., Seliya, N., Wald, R.,
	and Muharemagic, E. (2015).
	\newblock Deep learning applications and challenges in big data analytics.
	\newblock {\em Journal of Big Data}, 2(1):1.
	
	\bibitem[Neal, 1994]{neal1994priors}
	Neal, R.~M. (1994).
	\newblock Priors for infinite networks.
	\newblock {\em Technical Report}.
	
	\bibitem[Neal, 1996]{neal2012bayesian}
	Neal, R.~M. (1996).
	\newblock {\em Bayesian Learning for Neural Networks}, volume 118.
	\newblock Springer Science \& Business Media.
	
	\bibitem[Nychka et~al., 2015]{nychka2015multiresolution}
	Nychka, D., Bandyopadhyay, S., Hammerling, D., Lindgren, F., and Sain, S.
	(2015).
	\newblock A multiresolution {Gaussian} process model for the analysis of large
	spatial datasets.
	\newblock {\em Journal of Computational and Graphical Statistics},
	24(2):579--599.
	
	\bibitem[Paciorek and Schervish, 2004]{paciorek2004nonstationary}
	Paciorek, C.~J. and Schervish, M.~J. (2004).
	\newblock Nonstationary covariance functions for {Gaussian} process regression.
	\newblock In {\em Advances in Neural Information Processing Systems}, pages
	273--280.
	
	\bibitem[Peng et~al., 2009]{peng2009emergency}
	Peng, R.~D., Bell, M.~L., Geyh, A.~S., McDermott, A., Zeger, S.~L., Samet,
	J.~M., and Dominici, F. (2009).
	\newblock Emergency admissions for cardiovascular and respiratory diseases and
	the chemical composition of fine particle air pollution.
	\newblock {\em Environmental Health Perspectives}, 117(6):957--963.
	
	\bibitem[Peters et~al., 2001]{peters2001increased}
	Peters, A., Dockery, D.~W., Muller, J.~E., and Mittleman, M.~A. (2001).
	\newblock Increased particulate air pollution and the triggering of myocardial
	infarction.
	\newblock {\em Circulation}, 103(23):2810--2815.
	
	\bibitem[Porter et~al., 2015]{porter2015investigating}
	Porter, W.~C., Heald, C.~L., Cooley, D., and Russell, B. (2015).
	\newblock Investigating the observed sensitivities of air-quality extremes to
	meteorological drivers via quantile regression.
	\newblock {\em Atmospheric Chemistry and Physics}, 15(18):10349--10366.
	
	\bibitem[Reich et~al., 2011]{reich2011bayesian}
	Reich, B.~J., Fuentes, M., and Dunson, D.~B. (2011).
	\newblock Bayesian spatial quantile regression.
	\newblock {\em Journal of the American Statistical Association},
	106(493):6--20.
	
	\bibitem[Rimstad and Omre, 2014]{rimstad2014skew}
	Rimstad, K. and Omre, H. (2014).
	\newblock Skew-{Gaussian} random fields.
	\newblock {\em Spatial Statistics}, 10:43--62.
	
	\bibitem[Sampson et~al., 2013]{sampson2013regionalized}
	Sampson, P.~D., Richards, M., Szpiro, A.~A., Bergen, S., Sheppard, L., Larson,
	T.~V., and Kaufman, J.~D. (2013).
	\newblock {A regionalized national universal Kriging model using partial least
		squares regression for estimating annual PM2.5 concentrations in
		epidemiology}.
	\newblock {\em Atmospheric Environment}, 75:383--392.
	
	\bibitem[Stein, 2014]{stein2014limitations}
	Stein, M.~L. (2014).
	\newblock Limitations on low rank approximations for covariance matrices of
	spatial data.
	\newblock {\em Spatial Statistics}, 8:1--19.
	
	\bibitem[Vidakovic, 2009]{vidakovic2009statistical}
	Vidakovic, B. (2009).
	\newblock {\em Statistical Modeling by Wavelets}, volume 503.
	\newblock John Wiley \& Sons.
	
	\bibitem[Wahba, 1990]{wahba1990spline}
	Wahba, G. (1990).
	\newblock {\em Spline Models for Observational Data}, volume~59.
	\newblock SIAM.
	
	\bibitem[Waller and Gotway, 2004]{waller2004applied}
	Waller, L.~A. and Gotway, C.~A. (2004).
	\newblock {\em Applied Spatial Statistics for Public Health Data}, volume 368.
	\newblock John Wiley \& Sons.
	
	\bibitem[Whittle, 1954]{whittle1954stationary}
	Whittle, P. (1954).
	\newblock On stationary processes in the plane.
	\newblock {\em Biometrika}, pages 434--449.
	
	\bibitem[{World Health Organization}, 2013]{world2013health}
	{World Health Organization} (2013).
	\newblock Health effects of particulate matter.
	\newblock {\em Policy implications for countries in eastern Europe, Caucasus
		and central Asia}, 1(1):2--10.
	
	\bibitem[Xu and Genton, 2017]{xu2017tukey}
	Xu, G. and Genton, M.~G. (2017).
	\newblock Tukey g-and-h random fields.
	\newblock {\em Journal of the American Statistical Association},
	112(519):1236--1249.
	
\end{thebibliography}

\begin{thebibliography}{}
	
	\bibitem[Ba et~al., 2012]{ba2012composite}
	Ba, S., Joseph, V.~R., et~al. (2012).
	\newblock Composite gaussian process models for emulating expensive functions.
	\newblock {\em The Annals of Applied Statistics}, 6(4):1838--1860.
	
	\bibitem[Banerjee, 1973]{banerjee1973generalized}
	Banerjee, K.~S. (1973).
	\newblock Generalized inverse of matrices and its applications.
	\newblock {\em Technometrics}, 15(1):197--197.
	
	\bibitem[Blomqvist et~al., 2019]{blomqvist2019deep}
	Blomqvist, K., Kaski, S., and Heinonen, M. (2019).
	\newblock Deep convolutional gaussian processes.
	\newblock In {\em Joint European Conference on Machine Learning and Knowledge
		Discovery in Databases}, pages 582--597. Springer.
	
	\bibitem[Cs{\'a}ji, 2001]{csaji2001approximation}
	Cs{\'a}ji, B.~C. (2001).
	\newblock Approximation with artificial neural networks.
	\newblock {\em Faculty of Sciences, Etvs Lornd University, Hungary}, 24:48.
	
	\bibitem[Damianou and Lawrence, 2013]{damianou2013deep}
	Damianou, A. and Lawrence, N. (2013).
	\newblock Deep gaussian processes.
	\newblock In {\em Artificial Intelligence and Statistics}, pages 207--215.
	
	\bibitem[Diggle et~al., 2007]{diggle2007spatial}
	Diggle, P.~J., Thomson, M.~C., Christensen, O., Rowlingson, B., Obsomer, V.,
	Gardon, J., Wanji, S., Takougang, I., Enyong, P., Kamgno, J., et~al. (2007).
	\newblock Spatial modelling and the prediction of loa loa risk: decision making
	under uncertainty.
	\newblock {\em Annals of Tropical Medicine \& Parasitology}, 101(6):499--509.
	
	\bibitem[Freedman and Diaconis, 1981]{freedman1981histogram}
	Freedman, D. and Diaconis, P. (1981).
	\newblock On the histogram as a density estimator: {L}2 theory.
	\newblock {\em Zeitschrift f{\"u}r Wahrscheinlichkeitstheorie und verwandte
		Gebiete}, 57(4):453--476.
	
	\bibitem[Gal and Ghahramani, 2016]{gal2016dropout}
	Gal, Y. and Ghahramani, Z. (2016).
	\newblock Dropout as a bayesian approximation: Representing model uncertainty
	in deep learning.
	\newblock In {\em International Conference on Machine Learning}, pages
	1050--1059.
	
	\bibitem[{GPy}, 2012]{gpy2014}
	{GPy} (since 2012).
	\newblock {GPy}: A {Gaussian} process framework in python.
	\newblock \url{http://github.com/SheffieldML/GPy}.
	
	\bibitem[Ketkar et~al., 2017]{ketkar2017deep}
	Ketkar, N. et~al. (2017).
	\newblock {\em Deep Learning with Python}.
	\newblock Springer.
	
	\bibitem[Kingma and Ba, 2014]{kingma2014adam}
	Kingma, D.~P. and Ba, J. (2014).
	\newblock Adam: A method for stochastic optimization.
	\newblock {\em arXiv preprint arXiv:1412.6980}.
	
	\bibitem[Kumar et~al., 2018]{kumar2018deep}
	Kumar, V., Singh, V., Srijith, P., and Damianou, A. (2018).
	\newblock Deep gaussian processes with convolutional kernels.
	\newblock {\em arXiv preprint arXiv:1806.01655}.
	
	\bibitem[Lee et~al., 2017]{lee2017deep}
	Lee, J., Bahri, Y., Novak, R., Schoenholz, S.~S., Pennington, J., and
	Sohl-Dickstein, J. (2017).
	\newblock Deep neural networks as gaussian processes.
	\newblock {\em arXiv preprint arXiv:1711.00165}.
	
	\bibitem[Li et~al., 2019]{li2019deep}
	Li, R., Bondell, H.~D., and Reich, B.~J. (2019).
	\newblock Deep distribution regression.
	\newblock {\em arXiv preprint arXiv:1903.06023}.
	
	\bibitem[Neal, 1994]{neal1994priors}
	Neal, R.~M. (1994).
	\newblock Priors for infinite networks.
	\newblock {\em Technical Report}.
	
	\bibitem[Neal, 1996]{neal2012bayesian}
	Neal, R.~M. (1996).
	\newblock {\em Bayesian Learning for Neural Networks}, volume 118.
	\newblock Springer Science \& Business Media.
	
	\bibitem[Nychka et~al.(2015)]{Nychka:15}
	Nychka, D., Bandyopadhyay, S., Hammerling, D., Lindgren, F., and Sain, S. (2015). A multi-resolution {Gaussian} process model for the analysis of large spatial datasets. \textit{Journal of Computational and Graphical Statistics,} 24(2):579--599.
	
	\bibitem[Posch et~al., 2019]{posch2019variational}
	Posch, K., Steinbrener, J., and Pilz, J. (2019).
	\newblock Variational inference to measure model uncertainty in deep neural
	networks.
	\newblock {\em arXiv preprint arXiv:1902.10189}.
	
	\bibitem[Zammit-Mangion et~al., 2019]{zammit2019deep}
	Zammit-Mangion, A., Ng, T. L.~J., Vu, Q., and Filippone, M. (2019).
	\newblock Deep compositional spatial models.
	\newblock {\em arXiv preprint arXiv:1906.02840}.
	
\end{thebibliography}

\end{document}